\documentclass[journal]{IEEEtran}
\usepackage{amsmath,amsfonts}
\usepackage{algorithmic}
\usepackage{algorithm}
\usepackage{array}
\usepackage{textcomp}
\usepackage{stfloats}
\usepackage{url}
\usepackage{verbatim}
\usepackage{graphicx}
\usepackage{cite}
\usepackage{booktabs}
\usepackage{multirow}
\usepackage{multicol}
\usepackage{colortbl}
\usepackage{makecell}
\usepackage{xcolor}
\begin{document}

\title{Speech-FT: Merging Pre-trained And Fine-Tuned Speech Representation Models For Cross-Task Generalization}

\author{Tzu-Quan Lin, Wei-Ping Huang, Hao Tang, Hung-yi Lee}

\maketitle

\begin{abstract}
Fine-tuning speech representation models can enhance performance on specific tasks but often compromises their cross-task generalization ability. 
This degradation is often caused by excessive changes in the representations, making it difficult to retain information learned during pre-training. 
Existing approaches, such as regularizing weight changes during fine-tuning, may fail to maintain sufficiently high feature similarity with the pre-trained model, and thus could possibly lose cross-task generalization.
To address this issue, we propose Speech-FT, a novel two-stage fine-tuning framework designed to maintain cross-task generalization while benefiting from fine-tuning.
Speech-FT first applies fine-tuning specifically designed to reduce representational drift, followed by weight-space interpolation with the pre-trained model to restore cross-task generalization.
Extensive experiments on HuBERT, wav2vec 2.0, DeCoAR 2.0, and WavLM Base+ demonstrate that Speech-FT consistently improves performance across a wide range of supervised, unsupervised, and multitask fine-tuning scenarios.
Moreover, Speech-FT achieves superior cross-task generalization compared to fine-tuning baselines that explicitly constrain weight changes, such as weight-space regularization and LoRA fine-tuning.
Our analysis reveals that Speech-FT maintains higher feature similarity to the pre-trained model compared to alternative strategies, despite allowing larger weight-space updates.
Notably, Speech-FT achieves significant improvements on the SUPERB benchmark. For example, when fine-tuning HuBERT on automatic speech recognition, Speech-FT is able to reduce phone error rate from 5.17\% to 3.94\%, lower word error rate from 6.38\% to 5.75\%, and increase speaker identification accuracy from 81.86\% to 84.11\%. 
Speech-FT provides a simple yet powerful solution for further refining speech representation models after pre-training.
\end{abstract}


\section{Introduction}

\IEEEPARstart{S}{peech} representation models have become a cornerstone of modern speech processing~\cite{baevski2020wav2vec,ling2020decoar,hsu2021hubert,chen2022wavlm}, offering strong generalization across a wide range of downstream tasks such as phoneme recognition, speaker identification, and emotion recognition~\cite{yang2021superb}.
These models learn general speech representations from large-scale unlabeled data via self-supervised pre-training.
Recently, there has been growing interest in further enhancing these representations through fine-tuning, either using supervised signals~\cite{chen2021speech} or by continuing the self-supervised pre-training process~\cite{getman2024happens}.

Despite its potential benefits, fine-tuning often leads to a degradation in cross-task generalization ability~\cite{chen2021speech}. A fine-tuned model may become overly specialized to the target task, resulting in poor transferability to other tasks.
While this issue has been partially attributed to changes in the pre-trained features~\cite{kumar2022fine}, the underlying causes remain unclear. 
One common belief is that large deviations in weight space lead to representational drift~\cite{kirkpatrick2017overcoming, xu2019forget, chen-etal-2020-recall}. 
However, this is not necessarily the case. 
Even with substantial weight updates, a model can retain high feature similarity to the original representations, depending on how fine-tuning is conducted~\cite{raghavan2024engineering}.

A straightforward approach to preserving cross-task generalization is weight-space regularization, which penalizes large deviations from the pre-trained model during fine-tuning. 
While this constraint reduces deviations in weight space, it does not necessarily maintain feature similarity.
As a result, models trained with weight-space regularization could possibly lose the relatively general representations learned during pre-training, leading to degraded performance on unrelated tasks.

In this work, we investigate \emph{\textbf{how fine-tuning can enhance pre-trained speech representation models while preserving cross-task generalization ability}}. 
To achieve this, we propose \textbf{Speech-FT} (\textbf{Speech} \textbf{F}ine-\textbf{T}uning), a novel two-stage fine-tuning framework designed to maintain cross-task generalization while benefiting from fine-tuning.
In the first stage, Speech-FT applies a fine-tuning strategy specifically designed to reduce representational drift with the pre-trained model. We refer to this approach as \emph{stable fine-tuning} throughout this paper.
In the second stage, Speech-FT further restores feature similarity and enhances cross-task generalization through weight-space interpolation between the pre-trained and fine-tuned models. 
This interpolation step can be interpreted as a form of model merging~\cite{ilharco2022editing}, which has been shown to effectively integrate information from different models.
In the context of Speech-FT, interpolation combines task-specific information acquired during fine-tuning with the relatively general representations of the pre-trained model.

As we will analyze in Section~\ref{sec:property-analysis}, Speech-FT achieves higher feature similarity with the pre-trained model compared to weight-space regularization, despite allowing larger parameter updates during fine-tuning.
This aligns with prior work~\cite{raghavan2024engineering} which demonstrates the existence of low-impact subspaces in weight space where substantial weight changes induce minimal functional change;
our findings provide empirical evidence suggesting that such invariant subspaces exist in the context of speech representation models as well.

We conduct extensive experiments primarily on HuBERT~\cite{hsu2021hubert}, across a variety of supervised and unsupervised fine-tuning scenarios. In addition, we validate the generality of Speech-FT on other speech representation models including wav2vec 2.0~\cite{baevski2020wav2vec}, DeCoAR 2.0~\cite{ling2020decoar}, and WavLM Base+~\cite{chen2022wavlm}.
Our evaluations on the SUPERB benchmark~\cite{yang2021superb} reveal that Speech-FT consistently outperforms other baselines that constrain weight deviation during fine-tuning, such as weight-space regularization and LoRA fine-tuning~\cite{hu2022lora}.

In summary, the main contributions of this work are:
\begin{itemize}
    \item We propose \textbf{Speech-FT}, a novel two-stage fine-tuning framework for speech representation models that designed to maintain cross-task generalization while benefiting from fine-tuning.
    \item We evaluate Speech-FT across diverse supervised and unsupervised fine-tuning scenarios, using various speech representation models, demonstrating its generality.
    \item Speech-FT offers an efficient and effective approach to further improve general speech representations after pre-training.
    \item We conduct extensive comparisons against a variety of baselines, including regularization-based methods and LoRA fine-tuning, and show that Speech-FT consistently achieves stronger performance.
    \item We conduct an in-depth analysis showing that Speech-FT’s interpolation step plays a crucial role in restoring feature similarity with the pre-trained model, enabling it to maintain cross-task generalization ability while improving task-specific performance.
\end{itemize}

\section{Methodology}
\subsection{Problem Formulation}
\label{sec:problem-formulation}
Let the pre-trained speech representation model be denoted as $\theta_{0}$, whose representations are evaluated on a set of downstream tasks $\mathcal{T} = \{t_1, t_2, \dots, t_n\}$. 
Given a particular fine-tuning task $\hat{t}$, our objective is to design a fine-tuning algorithm that produces a representation model $\hat{\theta}$ which improves performance across tasks in $\mathcal{T}$.

\begin{figure*}[!t]
    \centering
    \includegraphics[width=1\linewidth]{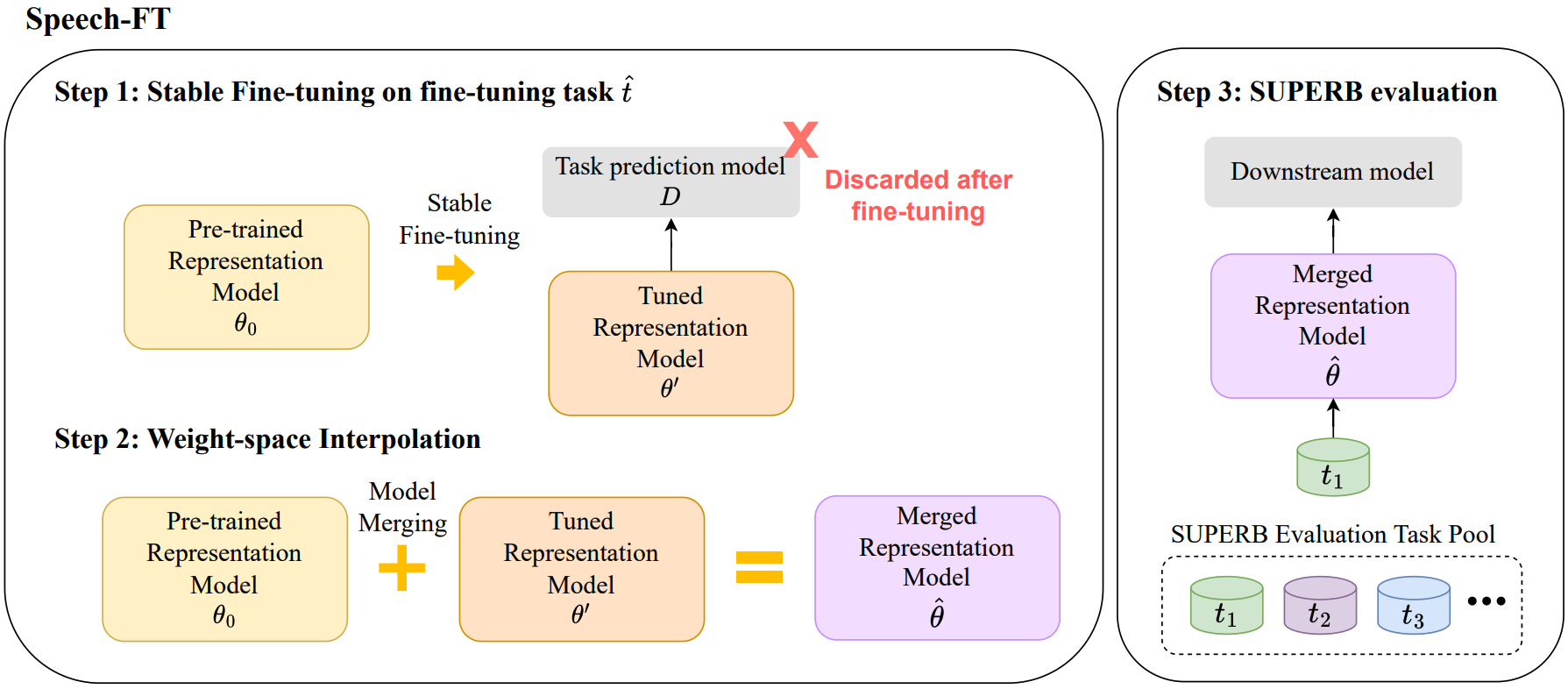}
    \caption{The pipeline of Speech-FT for representation learning and evaluation. 
    Step 1: A pre-trained representation model $\theta_0$ undergoes stable fine-tuning on a specific task $\hat{t}$, 
    producing a tuned representation model $\theta'$ while discarding the task prediction model $D$. 
    Step 2: The pre-trained and tuned models are merged in weight space to obtain the final representation model $\hat{\theta}$. 
    Step 3: The merged model is evaluated on the SUPERB benchmark by re-training task-specific downstream models, 
    ensuring that cross-task generalization of $\hat{\theta}$ is measured rather than performance of the discarded $D$.}
    \label{fig:pipeline}
\end{figure*}

During fine-tuning, we attach a task prediction model $D$ to the speech representation model. The task prediction model $D$ takes the speech representations as input and produces task-specific predictions. Both $D$ and $\theta_{0}$ are optimized jointly during fine-tuning. After fine-tuning, the tuned task prediction model $D$ is discarded (Step 1 in Figure~\ref{fig:pipeline}).
It is worth noting that we evaluate the cross-task performance of the representation model, rather than that of each tuned task prediction model $D$. To examine whether the representation model preserves its cross-task generalization ability, we adopt the SUPERB framework~\cite{yang2021superb} (Step 3 in Figure~\ref{fig:pipeline}), where downstream models are re-tuned individually without relying on the task prediction models trained during the initial fine-tuning stage.

\begin{algorithm}
    \caption{Speech-FT Procedure}
    \label{alg:speech-ft}
    \begin{algorithmic}[1]
        \STATE \textbf{Input:} Pre-trained model weights $\theta_0$, task prediction model $D$, fine-tuning task $\hat{t}$, interpolation scaling factor $\alpha$, stable fine-tuning rate $\beta$, total fine-tuning steps $S$
        \STATE \textbf{Output:} Fine-tuned model weights $\hat{\theta}$

        \STATE \textbf{Step 1: Freeze Downsampling Module}
        \STATE Load the pre-trained model with weights $\theta_0$
        \STATE Freeze the downsampling module (e.g., the CNN feature extractor)

        \STATE \textbf{Step 2: Stable Fine-tuning on Task $\hat{t}$}
        \FOR{step $t = 1$ to $S$}
            \IF{$t \leq \beta \% \cdot S$}
                \STATE Update only the task prediction model $D$
            \ELSE
                \STATE Update the entire model except the downsampling module
            \ENDIF
        \ENDFOR

        \STATE \textbf{Step 3: Weight-space Interpolation}
        \STATE Let $\theta'$ be the model weights after stable fine-tuning
        \STATE Compute the merged model weights using linear interpolation:
        \[
            \hat{\theta} = (1 - \alpha) \cdot \theta_0 + \alpha \cdot \theta'
        \]
        \vspace{-2em}
        \STATE \textbf{Return} $\hat{\theta}$
        
    \end{algorithmic}
\end{algorithm}

\subsection{Speech-FT}
Speech-FT consists of two main components: stable fine-tuning and weight-space interpolation.
Algorithm~\ref{alg:speech-ft} and Steps 1 and 2 of Figure~\ref{fig:pipeline} provide an overview of the Speech-FT procedure.
\label{sec:speech-ft}

\vspace{1em}
\noindent\textbf{Stable Fine-tuning (Stable-FT).}
Fine-tuning speech representation models is challenging, as 
pre-trained representations can undergo significant drift due to mismatches between pre-training and fine-tuning tasks.
To address this issue, we propose a stable fine-tuning process that helps minimize such drift.

When fine-tuning with a randomly initialized task prediction model $D$, the pre-trained features may be severely altered as the model adapts to the fine-tuning task~\cite{kumar2022fine}. This issue can be mitigated by first training a well-initialized task prediction model $D$. Therefore, we begin fine-tuning by updating only the task prediction model $D$ during the initial $\beta\%$ of fine-tuning steps.

We further stabilize fine-tuning by freezing the downsampling module.
Modern speech representation models often include a downsampling module at the beginning, typically a multi-layer convolutional encoder that processes raw audio and generates downsampled features.
Prior studies have shown that this module captures low-level features, such as frequency patterns~\cite{lin2023melhubert}, which are generally crucial for various applications.
To preserve these low-level features, we freeze the downsampling module throughout the entire fine-tuning process.

\vspace{1em}
\noindent\textbf{Weight-space Interpolation.}
While stable fine-tuning could reduce representational drift to some extent, it may still be insufficient for fully preserving the cross-task generalization ability of the pre-trained model.

A straightforward approach to mitigating this issue is to add a weight-space regularization term during fine-tuning to keep the model closer to the pre-trained parameters in weight space. 
While this method can reduce deviation in weight-space, it might fail to maintain feature similarity with the pre-trained model, thereby limiting its effectiveness in preserving relatively general representations.

Instead, we adopt model merging as an alternative, combining the pre-trained and fine-tuned models using linear interpolation.  
Model merging techniques have been shown to effectively integrate information from different models~\cite{ilharco2022editing, wortsman2022robust, lin2023spurious, lin2024mitigating, kulshreshtha2024sequential}. 
In our context, interpolating between the pre-trained and fine-tuned models leverages the pre-trained model’s strong cross-task generalization ability while still capturing task-specific improvements.
Notably, this approach also helps restore feature similarity with the pre-trained model—an aspect that weight-space regularization often fails to preserve.

Formally, the weight-space interpolation is defined as follows: 
Given the pre-trained model's weights $\theta_{0}$ and the fine-tuned model's weights $\theta'$, we derive the merged model $\hat{\theta}$ as follows:
\begin{equation}
\label{eq:interpolate}
\hat{\theta} = (1-\alpha) \cdot \theta_0 + \alpha \cdot \theta'
\end{equation}
where $\alpha$ is a scaling hyperparameter ranging from 0 to 1, determining the balance between the pre-trained and fine-tuned models.

\subsection{Scenarios with Multiple Fine-tuning Tasks}
\label{sec:multiple-task}
Speech-FT can also be applied to scenarios involving multiple fine-tuning tasks.
By leveraging multiple fine-tuning tasks $\hat{t}_1, \hat{t}_2, \dots, \hat{t}_k$, we aim to incorporate diverse information during fine-tuning, further enhancing speech representations.

We explore four different strategies for integrating multiple fine-tuning tasks:

\vspace{1em}
\noindent\textbf{Multitask Fine-tuning (MTF).}  
We perform stable fine-tuning on the pre-trained model using multiple tasks simultaneously, where separate task prediction models are appended to the pre-trained model. The resulting multitask model $\theta'$ is employed in Speech-FT as described in Equation~\ref{eq:interpolate}.

\noindent\textbf{Linear Merging.}  
We perform stable fine-tuning on a separate model for each task, resulting in $\theta'_1, \theta'_2,\dots, \theta'_k$. These models are then merged by averaging as follows:
\begin{equation}
\hat{\theta} = (1-\alpha) \cdot \theta_0 + \alpha \cdot \left(\frac{1}{k} \sum_{i=1}^{k} \theta_i'\right)
\end{equation}

\noindent\textbf{TIES Merging.}  
In contrast to linear merging, we adopt a more sophisticated merging technique known as TIES merging~\cite{yadav2024ties}.
TIES-Merging addresses parameter interference across fine-tuned models by performing three key steps: (i) trimming redundant parameter updates (those minimally changed during fine-tuning), (ii) resolving sign conflicts in parameter directions, and (iii) merging only the parameters whose sign agrees with the chosen consensus sign.  
The resulting model in Speech-FT is given by:
\begin{equation}
\hat{\theta} = (1-\alpha) \cdot \theta_0 + \alpha \cdot \text{TIES}(\theta_1', \theta_2', \dots, \theta_k')
\end{equation}

\noindent\textbf{Sequential Fine-tuning.}  
We apply Speech-FT to each fine-tuning task $\hat{t}_1, \hat{t}_2, \dots, \hat{t}_k$ in sequence, using the output model from the previous task as the pre-trained model for the next. This yields a final model that has been fine-tuned on all tasks one after another.

\section{Experiments}
All experiments are conducted on HuBERT~\cite{hsu2021hubert}, unless otherwise specified.
We evaluate the generality of our method on wav2vec 2.0~\cite{baevski2020wav2vec}, DeCoAR 2.0~\cite{ling2020decoar}, and WavLM Base+~\cite{chen2022wavlm} in Section~\ref{sec:generalizability}.
HuBERT, wav2vec 2.0, DeCoAR 2.0, and WavLM Base+ contain approximately 94M, 95M, 90M, and 95M parameters, respectively.
Instead of pre-training these models from scratch, we use publicly available pre-trained models\footnote{\url{https://github.com/facebookresearch/fairseq}}\footnote{\url{https://github.com/awslabs/speech-representations/}}\footnote{\url{https://github.com/microsoft/unilm/tree/master/wavlm}} as the starting point for fine-tuning.
While fine-tuning inherently requires additional task-specific labeled data beyond the pre-training stage, all fine-tuning methods in this work are trained on exactly the same datasets and splits, ensuring fully fair comparisons.

For each fine-tuning task, we select the checkpoint with the best validation score.
For weight-space interpolation, we fix the scaling factor $\alpha=0.25$ throughout the paper and discuss its impact in Section~\ref{sec:property-analysis}. We observe that $\alpha=0.25$ generalizes well across different fine-tuning tasks.
For stable fine-tuning, where only the task prediction model $D$ is updated during the initial $\beta\%$ of fine-tuning steps, we set $\beta=10\%$ across all experiments, as preliminary experiments showed that a short warm-up is sufficient.
We fine-tune all model parameters except the downsampling module, which is kept frozen for stable fine-tuning. 
In Section~\ref{sec:strong-baselines}, we additionally include parameter-efficient fine-tuning baselines for comparison.
All fine-tuning experiments are performed on a single NVIDIA RTX 3090 GPU.
We outline the hyperparameters, objectives, datasets, task prediction models $D$, and computational budget used for different fine-tuning tasks below.

\subsection{Supervised Fine-tuning Tasks}
To ensure a balanced and representative evaluation, we select two tasks from each of four major categories commonly studied in speech processing: recognition-related tasks, phoneme-related tasks, speaker-related tasks, and emotion-related tasks.

\vspace{1em}
\noindent\textbf{Automatic Speech Recognition (ASR) on TED-LIUM~\cite{rousseau2012ted}.}
The task prediction model is a three-layer LSTM trained using CTC loss. Fine-tuning is conducted for 50,000 steps with an effective batch size of 32 and a learning rate of $10^{-4}$. We limit the training set to a randomly selected 100-hour subset. The entire process takes approximately 12 hours.

\vspace{1em}
\noindent\textbf{Phoneme Classification (PC) on TIMIT~\cite{garofolo1993darpa}.}
The task prediction model is a linear projection layer trained with cross-entropy loss. Fine-tuning is performed for 300,000 steps with a batch size of 16 and a learning rate of $10^{-4}$. Following prior work~\cite{lopes2011phone}, we use standard training and validation splits. The entire process takes about 6 hours.

\vspace{1em}
\noindent\textbf{Speaker Identification (SID) on Librispeech~\cite{panayotov2015librispeech}.} \\
The task prediction model is a linear projection layer that takes mean-pooled speech representations as input and is trained with cross-entropy loss. Fine-tuning is conducted for 100,000 steps using a batch size of 32 and a learning rate of $2 \times 10^{-4}$. Following prior work~\cite{oord2018representation}, we randomly sample a 10-hour subset while maintaining the same number of speakers. The entire process takes around 12 hours.

\vspace{1em}
\noindent\textbf{Emotion Recognition (ER) on CREMA-D~\cite{cao2014crema}.}
CREMA-D is an audio-visual emotional dataset.
Only speech-based emotional annotations
are used.
The task prediction model is a linear projection layer that takes mean-pooled speech representations as input and is trained with cross-entropy loss. Fine-tuning is performed for 15,510 steps with a batch size of 32 and a learning rate of $10^{-4}$. Following the EMO-SUPERB setup~\cite{wu2024emo}, the training set contains approximately 6.5 hours of speech, and the entire process takes around 5 hours.

\vspace{1em}
\noindent\textbf{Automatic Speech Recognition (ASR) on Librispeech.}
The task prediction model is a three-layer LSTM trained with CTC loss. Fine-tuning is conducted for 200,000 steps with a batch size of 32 and a learning rate of $10^{-4}$, using the 100-hour subset of Librispeech for training and the dev-clean subset for validation. The entire process takes approximately 28 hours.

\vspace{1em}
\noindent\textbf{Phoneme Recognition (PR) on Librispeech.}
The task prediction model is a linear projection layer trained with CTC loss. Fine-tuning is performed for 100,000 steps with a batch size of 32 and a learning rate of $10^{-4}$, using the 100-hour subset of Librispeech for training and the dev-clean subset for validation. The entire process takes approximately 15 hours.

\vspace{1em}
\noindent\textbf{Speaker Identification (SID) on VoxCeleb1~\cite{nagrani2020voxceleb}.}
The task prediction model is a linear projection layer that processes mean-pooled speech representations and is trained with cross-entropy loss. Fine-tuning is conducted for 200,000 steps with a batch size of 32 and a learning rate of $10^{-4}$, using approximately 318 hours of speech data. The entire process takes around 30 hours.

\vspace{1em}
\noindent\textbf{Emotion Recognition (ER) on IEMOCAP~\cite{busso2008iemocap}.}
The task prediction model is a linear projection layer that processes mean-pooled speech representations and is trained with cross-entropy loss. Fine-tuning is conducted for 30,000 steps with a batch size of 32 and a learning rate of $10^{-4}$, using Sections 2 to 5 for training and Section 1 for validation. The training set contains approximately 5.56 hours of speech, and the entire process takes around 5 hours.

\subsection{Unsupervised Fine-tuning Tasks}
\vspace{1em}
For unsupervised fine-tuning, we perform continued pre-training of HuBERT on AISHELL-3~\cite{shi2020aishell}, a widely used Mandarin speech corpus published by Beijing Shell Shell Technology Co.,Ltd. 
Our goal is to examine whether Speech-FT enables the model to adapt to a new language (Chinese) while maintaining its performance on a previously learned language (English).
We perform continued pre-training for 100,000 steps with an effective batch size of 32 and a learning rate of $10^{-5}$.
The entire process takes approximately 24 hours.
Since the data in AISHELL-3 has a sampling rate of 44100, we resample it to 16000 to match the sampling rate of HuBERT.
The training dataset of AISHELL-3 consist of approximately 63.17 hours of data.

Since the k-means model used to generate the targets for HuBERT Base is not publicly available, we follow prior work~\cite{lin2024daisy} and use HuBERT Base’s own predictions (via $\arg\max$ over codeword logits) as pseudo targets for continued pre-training.
We then optimize a simplified HuBERT loss~\cite{lin2023melhubert, yang2023fast} using cross-entropy over masked time steps, which avoids the need for a learnable codebook and offers stable and efficient training.
We refer to this method as continued pre-training, as it extends the pre-training process of the original model. From another perspective, it can also be regarded as a form of self-training~\cite{kahn2020self}, since the pseudo targets are generated using the model itself.

\begin{table*}[ht!]
    \centering
    \caption{SUPERB downstream results for eight supervised fine-tuning tasks. Background color highlights cases where the fine-tuning task matches the evaluation task.}
    \begin{tabular}{l l c c c cc c}
        \toprule
        \multirow{3}{*}{\textbf{Fine-tuning Task}} & \multirow{3}{*}{\textbf{Method}} & \textbf{PR on} & \textbf{SID on} & \textbf{ER on} & \multicolumn{2}{c}{\textbf{SF on}} & \multirow{3}{*}{\textbf{$\text{SUPERB}_{\text{S}}\uparrow$}}\\
        & & \textbf{Librispeech} & \textbf{Voxceleb1} & \textbf{IEMOCAP} & \multicolumn{2}{c}{\textbf{SNIPS}} \\
        \cmidrule(r){3-7}
        & & \textbf{PER\%$\downarrow$} & \textbf{ACC\%$\uparrow$} & \textbf{ACC\%$\uparrow$} & \textbf{F1$\uparrow$} & \textbf{CER\%$\downarrow$} & \\
        \midrule
        \multirow{2}{*}{ASR on TED-LIUM} & Speech-FT & 3.94 & 84.11 & 67.78 & 88.84 & 23.46 & 905.79\\
        & Stable-FT & 5.22 & 79.89 & 65.38 & 88.52 & 23.80 & 870.71\\
        \midrule
        \multirow{2}{*}{PC on TIMIT} & Speech-FT & 4.76 & 81.78 & 65.48 & 88.65 & 24.05 & 877.66\\
        & Stable-FT & 10.34 & 66.34 & 61.25 & 83.50 & 33.82 & 726.64\\
        \midrule
        \multirow{2}{*}{SID on Librispeech} & Speech-FT & 5.67 & 83.24 & 65.48 & 87.98 & 24.90 & 877.97\\
        & Stable-FT & 14.18 & 75.52 & 59.51 & 84.73 & 31.54 & 847.33\\
        \midrule
        \multirow{2}{*}{ER on CREMA-D} & Speech-FT & 5.02 & 83.36 & 65.59 & 88.41 & 24.75 & 871.93\\
        & Stable-FT & 6.62 & 81.48 & 64.69 & 87.28 & 27.03 & 741.19\\
        \midrule
        \multirow{2}{*}{ASR on Librispeech} & Speech-FT & 3.93 & 82.79 & 66.38 & 87.97 & 24.93 & 882.44\\
        & Stable-FT & 5.35 & 72.14 & 63.82 & 86.01 & 29.32 & 805.33\\
        \midrule
        \multirow{2}{*}{PR on Librispeech} & Speech-FT & \cellcolor{gray!20} 2.99 & 81.95 & 65.79 & 88.60 & 24.52 & 883.93\\
        & Stable-FT & \cellcolor{gray!20} 2.21 & 75.60 & 63.14 & 88.09 & 26.24 & 842.03\\
        \midrule
        \multirow{2}{*}{SID on Voxceleb1} & Speech-FT & 6.91 & \cellcolor{gray!20} 88.80 & 66.03 & 87.20 & 25.23 & 881.07\\
        & Stable-FT & 29.08 & \cellcolor{gray!20} 87.75 & 60.29 & 77.64 & 43.27 & 651.42\\
        \midrule
        \multirow{2}{*}{ER on IEMOCAP} & Speech-FT & 4.95 & 82.63 & \cellcolor{gray!20} 85.95 & 87.23 & 25.80 & 1010.29\\
        & Stable-FT & 16.11 & 73.07 & \cellcolor{gray!20} 90.07 & 81.40 & 36.62 & 908.86\\
        \midrule
        \midrule
        \multicolumn{2}{c}{Pre-trained} & 5.17 & 81.86 & 64.99 & 88.54 & 24.70 & 870.20\\
        \bottomrule
    \end{tabular}
    \label{tab:supervised-finetune}
\end{table*}

\subsection{Evaluation}
\noindent\textbf{Datasets and tasks.}
To evaluate the cross-task generalization ability of speech representations, we use the SUPERB~\cite{yang2021superb} benchmark, which assesses models across a diverse set of downstream tasks, covering content, speaker, paralinguistic, and semantic aspects.
Evaluating the entire SUPERB benchmark is computationally expensive due to its wide range of tasks. 
MiniSUPERB~\cite{wang2023minisuperb} addresses this by selecting representative tasks, demonstrating that evaluation on a well-chosen subset sufficiently reflects a model’s performance.
Following this approach, we evaluate four tasks: phoneme recognition (PR, content) on Librispeech~\cite{panayotov2015librispeech}, speaker identification (SID, speaker) on Voxceleb1~\cite{nagrani2020voxceleb}, emotion recognition (ER, paralinguistic) on IEMOCAP~\cite{busso2008iemocap}, and slot filling (SF, semantic) on Audio SNIP \cite{lai2021semi}.
To provide a more comprehensive evaluation, we additionally assess full SUPERB benchmark performance on representative models in Section~\ref{sec:generalizability}.
We report phone error rate (PER\%) for PR, accuracy (ACC\%) for SID and ER, and slot-type F1 score (F1) along with slot value CER (CER\%) for SF.
We adhere to the default SUPERB evaluation setup, where the speech representation model remains frozen, and the weighted sum of its layer-wise features is used as input for the downstream model. 
Recall from Section~\ref{sec:problem-formulation} that our goal is to evaluate cross-task generalization. As such, the task prediction model $D$ is discarded after fine-tuning. Instead, we train a separate downstream model for each SUPERB evaluation task.
During training, only the learnable weighted sum and the downstream model are updated. 
To ensure the stability of the weighted sum features, we apply layer normalization to the representations before computing the sum. The performance of HuBERT, wav2vec 2.0, DeCoAR 2.0, and WavLM Base+ reported in this paper has been reproduced with this modification.
We adhere to the default number of update steps specified by the official SUPERB evaluation tool, s3prl\footnote{\url{https://github.com/s3prl/s3prl/tree/main}}, for downstream model training. 

\vspace{1em}
\noindent\textbf{SUPERB Score.} 
To facilitate a more intuitive and unified comparison of different speech representation models, we adopt the SUPERB score ($\text{SUPERB}_{\text{S}}$) as introduced in the SUPERB SLT 2022 Challenge~\cite{feng2023superb}. This score linearly scales each task's performance between a traditional baseline (FBank) and the best known result (SOTA) from prior work, reflecting how much each model improves upon traditional features relative to state-of-the-art representations. When traditional spectral features already perform close to SOTA, even small improvements are treated as more significant—implicitly accounting for task difficulty.
For tasks with multiple evaluation metrics, an average is first computed across metrics within the same task, followed by an average across all tasks. Mathematically, let $\phi_{t,j}(f)$ be the score of model $f$ on metric $j$ of task $t$, with $\phi_{t,j}(\text{Baseline})$ and $\phi_{t,j}(\text{SOTA})$ denoting the corresponding baseline and SOTA values, respectively. Define $\mathcal{T}$ as the set of evaluation tasks and $\mathcal{M}_t$ as the set of metrics for task $t$. The full formulation is:

\begin{equation} 
\Phi_{t,j}(f) = \frac{\phi_{t,j}(f) - \phi_{t,j}(\text{Baseline})}{\phi_{t,j}(\text{SOTA}) - \phi_{t,j}(\text{Baseline})} \end{equation} \begin{equation} 
\text{SUPERB}_\text{S}(f) = \frac{1000}{|\mathcal{T}|} \sum_{t \in \mathcal{T}}
\frac{1}{|\mathcal{M}_t|} \sum_{j \in \mathcal{M}_t} \Phi_{t,j}(f) 
\label{eq:superb-score}
\end{equation}

The performance of both FBank and SOTA models is obtained from the SUPERB journal extension~\cite{yang2024large}. The final score is scaled by a factor of 1000 for readability. In this paper, we report $\text{SUPERB}_{\text{S}}$ based on four downstream tasks: PR, SID, ER, and SF. This metric has been widely adopted in previous studies for evaluating generalization across speech tasks~\cite{shi2023ml, shi2023multi, wang2023task, chang2024colld}.

\begin{table}[t!] 
\centering 
\caption{
Results of Speech-FT on the unsupervised fine-tuning tasks, where we continue HuBERT pre-training on the Chinese speech corpus AISHELL-3 \cite{shi2020aishell}, the Spanish subset of the Multilingual LibriSpeech dataset~\cite{pratap2020mls}, and the English dataset VoxCeleb1~\cite{nagrani2020voxceleb}. Stable-FT (20\%) denotes the snapshot of the Stable-FT model saved at 20\% of the total fine-tuning steps.
}
\scalebox{0.88}{ 
\begin{tabular}{l l c c c} 
\toprule 
\multirow{3}{*}{\textbf{Language}} & \multirow{3}{*}{\textbf{Method}} & \multirow{2}{*}{\textbf{English}} & \textbf{Chinese} & \textbf{Spanish} \\ 
& & & \textbf{zh-CN ASR} & \textbf{es ASR} \\ 
\cmidrule(lr){3-5} 
& & \textbf{$\text{SUPERB}_{\text{S}}\uparrow$} & \textbf{CER\%$\downarrow$} & \textbf{WER\%$\downarrow$} \\ 
\cmidrule(lr){1-5} 
\multirow{3}{*}{\makecell[l]{Chinese\\(cross-lingual)}} & Speech-FT & 866.51 & 24.23 & - \\ 
& Stable-FT & 789.88 & 23.47 & - \\ 
& Stable-FT  (20\%) & 832.37 & 24.26 & - \\
\midrule
\multirow{2}{*}{\makecell[l]{Spanish\\(cross-lingual)}} & Speech-FT & 868.26 & - & 32.87 \\ 
& Stable-FT & 747.50 & - & 30.36 \\ 
\midrule
\multirow{2}{*}{\makecell[l]{English\\(mono-lingual)}} & Speech-FT & 876.10 & - & - \\ 
& Stable-FT & 867.96 & - & - \\ 
\midrule 
\midrule 
\multicolumn{2}{c}{Pre-trained} & 870.20 & 24.94 & 35.39 \\ 
\bottomrule 
\end{tabular} 
} 
\label{tab:continuous-chinese-unsupervised} 
\end{table}
\begin{table*}[ht]
\centering
\caption{Results of Speech-FT when using two fine-tuning tasks: phoneme recognition (PR) on Librispeech and speaker identification (SID) on VoxCeleb1. The background color highlights cases where the evaluation task is included in the fine-tuning tasks.}
\scalebox{1.0}{
\begin{tabular}{l l c c c cc c }
\toprule
\multirow{2}{*}{\textbf{Strategy}} & \multirow{2}{*}{\textbf{Method}} & \textbf{PR} & \textbf{SID} & \textbf{ER} & \multicolumn{2}{c}{\textbf{SF}} & \multirow{2}{*}{\textbf{$\text{SUPERB}_{\text{S}}\uparrow$}} \\
\cmidrule(r){3-7}
& & \textbf{PER\%$\downarrow$} & \textbf{ACC\%$\uparrow$} & \textbf{ACC\%$\uparrow$} & \textbf{F1$\uparrow$} & \textbf{CER\%$\downarrow$} & \\
\midrule
\multirow{2}{*}{MTF} & Speech-FT & \cellcolor{gray!20}4.27 & \cellcolor{gray!20}85.47 & 65.71 & 87.76 & 25.46 & 880.63 \\
& Stable-FT & \cellcolor{gray!20} 2.86 & \cellcolor{gray!20}86.57 & 57.06 & 83.95 & 31.07 & 785.85 \\
\midrule
\multirow{2}{*}{Linear Merge} & Speech-FT & \cellcolor{gray!20}4.32 & \cellcolor{gray!20}85.60 & 66.17 & 88.22 & 25.53 & 886.36 \\
& Stable-FT & \cellcolor{gray!20}9.80 & \cellcolor{gray!20}87.21 & 63.72 & 85.34 & 30.45 & 822.72 \\
\midrule
\multirow{2}{*}{TIES Merge} & Speech-FT & \cellcolor{gray!20}4.67 & \cellcolor{gray!20}86.23 & 65.46 & 88.47 & 24.77 & 885.95 \\
& Stable-FT & \cellcolor{gray!20}14.67 & \cellcolor{gray!20}88.57 & 62.46 & 84.24 & 32.45 & 788.91 \\
\midrule
\multirow{2}{*}{Sequential} & Speech-FT & \cellcolor{gray!20}4.77 & \cellcolor{gray!20} 87.61 & 65.80 & 88.15 & 25.45 & 887.50 \\
 & Stable-FT & \cellcolor{gray!20}25.72 & \cellcolor{gray!20}85.97 & 60.03 & 76.13 & 42.76 & 648.96 \\
\midrule
\midrule
\multicolumn{2}{c}{Pre-trained} & 5.17 & 81.86 & 64.99 & 88.54 & 24.70 & 870.20 \\
\bottomrule
\end{tabular}
}
\label{tab:multitask}
\end{table*}
\begin{table}[ht]
\centering
\caption{The results of Speech-FT under a gradually increasing number of fine-tuning tasks. ``PR'' denotes PR on Librispeech, ``SID'' denotes SID on Voxceleb1, and ``ER'' denotes ER on IEMOCAP.}
\scalebox{1.0}{
\begin{tabular}{l c c c cc c}
\toprule
\textbf{Fine-tuning Tasks} & \textbf{$\text{SUPERB}_{\text{S}}\uparrow$} \\
\midrule
PR  & 883.93 \\
SID & 881.07 \\
PR+SID & 886.36 \\
PR+SID+ER & 916.8 \\
\midrule
Pre-trained & 870.2 \\
\bottomrule
\end{tabular}
}
\label{tab:multitask-compare-single-simplied}
\end{table}

\vspace{1em}
\noindent\textbf{Baselines.}
Our goal is to improve pre-trained speech representation models through fine-tuning. Therefore, the most fundamental baseline is the pre-trained model itself, denoted as \textbf{Pre-trained}. 
Another key baseline is stable fine-tuning, abbreviated as \textbf{Stable-FT}.
We exclude regular fine-tuning from our baselines for the following reasons:
First, we empirically found that in some cases, regular fine-tuning fails entirely, preventing successful fine-tuning of the speech representation model.
Second, even in cases where regular fine-tuning succeeds, it consistently underperforms compared to Stable-FT, making Stable-FT a stronger and more reliable baseline\footnote{For example, on speaker identification (Voxceleb1), Stable-FT achieves a SUPERB score of 651.42 versus 568.62 with regular fine-tuning, and on emotion recognition (CREMA-D), Stable-FT achieves 847.33 versus 834.38.}.

The baselines can also be viewed as variations of $\alpha$ in Equation~\ref{eq:interpolate}, where \textbf{Stable-FT} corresponds to \textbf{Speech-FT} with $\alpha = 1.0$, and \textbf{Pre-trained} corresponds to \textbf{Speech-FT} with $\alpha = 0.0$.

In addition to these two core comparisons, we further benchmark Speech-FT against several fine-tuning strategies that constrain weight deviation during fine-tuning, as detailed in Section~\ref{sec:strong-baselines}. These strategies include:
(1) \textbf{Weight-space regularization}, which penalizes deviations from the pre-trained weights during fine-tuning;
(2) \textbf{Parameter-efficient fine-tuning}~\cite{hu2022lora, liu2024dora}, which inserts trainable lightweight adapters instead of modifying the full set of parameters;
(3) \textbf{Early stopping}, where checkpoints are taken early in the fine-tuning process to limit weight deviation.

\section{Main Results}
In this section, we show the effectiveness of Speech-FT in three different scenarios: supervised, unsupervised, and multitask fine-tuning.
\label{sec:main-results}

\begin{table}[ht]
\centering
\caption{Results of Speech-FT on wav2vec 2.0, DeCoAR 2.0, and WavLM Base+ when fine-tuning with ASR on TED-LIUM.}
\scalebox{1.0}{
\begin{tabular}{l l c }
\toprule
\textbf{Model} & \textbf{Method} & \textbf{$\text{SUPERB}_{\text{S}}\uparrow$} \\
\midrule
\multirow{3}{*}{wav2vec 2.0} & Speech-FT & \textbf{864.79} \\
& Stable-FT & 824.82 \\
& Pre-trained & 837.08 \\
\midrule
\multirow{3}{*}{DeCoAR 2.0} & Speech-FT & \textbf{781.77} \\
& Stable-FT & 701.60 \\
& Pre-trained & 736.77 \\
\midrule
\multirow{3}{*}{WavLM Base+} & Speech-FT & \textbf{945.50} \\
 & Stable-FT & 907.31 \\
& Pre-trained & 942.96 \\
\bottomrule
\end{tabular}
}
\label{tab:other-ssl}
\end{table}
\begin{table*}[ht]
\centering
\caption{The full SUPERB benchmark evaluation results for fine-tuning HuBERT and WavLM Base+ with ASR on TED-LIUM.  
Note that the SUPERB score ``$\text{SUPERB}_{\text{full}}$'' reported here is computed across the entire SUPERB benchmark and,  
therefore, is not directly comparable to the $\text{SUPERB}_{\text{s}}$ scores in this paper,  
which are based on only four evaluation tasks.}
\scalebox{0.95}{
\renewcommand{\arraystretch}{1.0}
\setlength{\tabcolsep}{2pt}
\begin{tabular}{l l c c c c c c c c c cc c}
\toprule
\multirow{2}{*}{\textbf{Model}} & \multirow{2}{*}{\textbf{Method}} & \textbf{PR} & \textbf{ASR} & \textbf{KS} & \textbf{QbE} & \textbf{SID} & \textbf{ASV} & \textbf{SD} & \textbf{ER} & \textbf{IC} & \multicolumn{2}{c}{\textbf{SF}} & \multirow{2}{*}{\textbf{$\text{SUPERB}_{\text{FULL}}\uparrow$}} \\
\cmidrule(r){3-13}
& & \textbf{PER\%$\downarrow$} & \textbf{WER\%$\downarrow$} & \textbf{ACC\%$\uparrow$} & \textbf{MTWV$\uparrow$} & \textbf{ACC\%$\uparrow$} & \textbf{EER\%$\downarrow$} & \textbf{DER\%$\downarrow$} & \textbf{ACC\%$\uparrow$} & \textbf{ACC\%$\uparrow$} & \textbf{F1$\uparrow$} & \textbf{CER\%$\downarrow$} & \\
\midrule
\multirow{3}{*}{HuBERT} & Speech-FT & \textbf{3.94} & \textbf{5.75} & 96.69 & \textbf{0.1011} & \textbf{84.11} & \textbf{5.27} & \textbf{5.99} & \textbf{67.78} & \textbf{98.97} & \textbf{88.84} & \textbf{23.46} & \textbf{874.51} \\
& Stable-FT & 5.22 & 7.63 & \textbf{96.92} & 0.0970 & 79.89 & 5.34 & 6.02 & 65.38 & 98.71 & 88.52 & 23.80 & 845.00 \\
& Pre-trained  & 5.17 & 6.38 & 96.59 & 0.0687 & 81.86 & 5.56 & 6.32 & 64.99 & 98.02 & 88.54 & 24.70 & 815.47 \\
\midrule
\multirow{3}{*}{WavLM Base+} & Speech-FT & \textbf{3.42} & \textbf{5.18} & \textbf{97.44} & 0.1090 & 88.12 & 4.34 & 3.84 & \textbf{69.02} & \textbf{99.10} & \textbf{90.73} & \textbf{21.10} & \textbf{950.60} \\
& Stable-FT & 3.93 & 6.29 & 96.72 & \textbf{0.1189} & 83.20 & 5.04 & 5.19 & 67.63 & 98.79 & 89.41 & 22.95 & 905.09 \\
& Pre-trained  & 3.94 & 5.57 & 97.36 & 0.0989 & \textbf{89.43} & \textbf{4.12} & \textbf{3.50} & 68.67 & 98.99 & 90.52 & 21.30 & 946.95 \\
\bottomrule
\end{tabular}
}
\label{tab:full-superb}
\end{table*}

\subsection{Supervised Fine-tuning}
We begin by fine-tuning HuBERT on various supervised tasks, including automatic speech recognition (ASR) on TED-LIUM~\cite{rousseau2012ted}, phoneme classification (PC) on TIMIT~\cite{garofolo1993darpa}, speaker identification (SID) on Librispeech~\cite{panayotov2015librispeech}, emotion recognition (ER) on CREMA-D~\cite{cao2014crema}, ASR on Librispeech, phoneme recognition (PR) on Librispeech, SID on VoxCeleb1~\cite{nagrani2020voxceleb}, and ER on IEMOCAP~\cite{busso2008iemocap}.

Table~\ref{tab:supervised-finetune} presents the SUPERB score results, where Speech-FT consistently outperforms Stable-FT across all supervised fine-tuning tasks. Additionally, Speech-FT improves performance compared to the pre-trained model for all tasks.
The table further provides a detailed breakdown of the downstream results, with the background color highlighting cases in which the fine-tuning task matches the evaluation task.
Overall, Stable-FT significantly degrades performance on unrelated tasks.  
For instance, fine-tuning on SID with VoxCeleb1 causes substantial degradation in PR and SF, with PER increasing by 23.91\% and CER by 18.57\%, respectively.  
Similarly, fine-tuning on SID with Librispeech leads to performance drops in PR and SF, with PER increasing by 9.01\% and CER by 6.84\%.  
Fine-tuning on PC with TIMIT negatively impacts SID, reducing ACC by 15.52\%.  

In contrast, Speech-FT not only preserves the cross-task generalization ability of the pre-trained model but also enhances overall representation quality, as reflected in the increased SUPERB score. Notably, fine-tuning on ASR tasks such as TED-LIUM yields consistent improvements across all evaluation tasks. We hypothesize that this is because ASR involves learning both phonetic and temporal patterns, which benefit other tasks through shared feature representations. For example, Speech-FT on TED-LIUM leads to a 1.23\% PER reduction on PR, a 2.25\% ACC increase on SID, a 2.79\% ACC increase on ER, and a 1.24\% CER reduction on SF. These results indicate that Speech-FT effectively leverages rich ASR supervision to improve speech representations. It is also worth noting that fine-tuning on IEMOCAP achieves the highest SUPERB score (1010.29), mainly due to the large improvement on ER (from 64.99\% to 85.95\%). For PR, SID, and SF, fine-tuning on ASR offers substantially greater benefits.

\subsection{Unsupervised Fine-tuning}

We further evaluate the effectiveness of Speech-FT in an unsupervised fine-tuning scenario, where no task-specific labels are available.
Specifically, we simulate a cross-lingual setting by continuing HuBERT pre-training on AISHELL-3~\cite{shi2020aishell}, a large-scale Mandarin speech corpus.
This setting mimics the realistic case where practitioners adapt pre-trained speech models to new domains or languages using only unlabeled data.

To evaluate whether the model retains its ability to represent the original language, we assess performance on English downstream tasks from SUPERB. 
Additionally, to examine adaptation to the new language, we introduce Mandarin ASR from the Common Voice corpus~\cite{ardila2019common} as an auxiliary evaluation task.

Results are shown in Table~\ref{tab:continuous-chinese-unsupervised}. 
We observe that continued pre-training alone (i.e., Stable-FT) severely degrades generalization to English tasks, causing a drop of 80.32 in the SUPERB score. 
In contrast, Speech-FT achieves a balanced trade-off: it improves performance on the Chinese ASR task while largely preserving the model’s English ability.
Compared with an earlier Stable-FT checkpoint (20\% of fine-tuning steps), Speech-FT yields similar Chinese ASR results but maintains substantially higher English performance, further highlighting its advantage.
This demonstrates that Speech-FT mitigates the representational forgetting typically observed in naive unsupervised adaptation and enables more effective cross-lingual fine-tuning.

In addition to Chinese, we also extend the setting to Spanish by continuing HuBERT pre-training on the Spanish subset of the Multilingual LibriSpeech dataset~\cite{pratap2020mls} and evaluating Spanish ASR on the Common Voice corpus~\cite{ardila2019common}. 
For Spanish, Speech-FT works better than in the Chinese setting, likely because Spanish is closer to the pretrained language, which is English.
Furthermore, we conduct mono-lingual unsupervised fine-tuning on VoxCeleb1, which contains in-the-wild interview speech with diverse speakers and noisy conditions. The results show that Speech-FT also improves the representation quality in this setting.

\subsection{Multiple Fine-tuning Tasks}
\label{sec:multiple-finetune}
We further evaluate Speech-FT in the context of multiple fine-tuning tasks. 
We first present the results using two fine-tuning tasks: PR on Librispeech and SID on VoxCeleb1.
Recall that there are four strategies for incorporating multiple fine-tuning tasks: multitask fine-tuning (MTF), linear merging, TIES merging, and sequential fine-tuning.
For sequential fine-tuning, we follow a fixed order where PR on Librispeech is applied first, followed by SID on VoxCeleb1.

The results are shown in Table~\ref{tab:multitask}.
Overall, Stable-FT significantly degrades performance on PR, ER, and SF.  
For example, under the sequential fine-tuning strategy, PR increases by 20.55\% PER, ER decreases by 4.96\% ACC, and SF increases by 18.06\% CER.  
Similarly, with the TIES merging strategy, PR increases by 9.5\% PER, ER decreases by 2.53\% ACC, and SF increases by 7.75\% CER.  
In contrast, Speech-FT effectively mitigates this degradation, leading to an overall improvement in the SUPERB score compared to the pre-trained model.

Interestingly, sequential fine-tuning performs the worst among the strategies when using Stable-FT alone. However, when Speech-FT is applied after each task, it becomes the best-performing strategy overall. We attribute this to the effect of repeated interpolation steps, which serve to gradually integrate task-specific information while continually restoring relatively general representations.
This process mitigates interference between task-specific representations and leads to improved cross-task generalization.

To further illustrate the benefits of Speech-FT in multiple fine-tuning tasks, we present the results of gradually increasing the number of fine-tuning tasks in Table~\ref{tab:multitask-compare-single-simplied}. 
In this experiment, we adopt linear merging as a representative strategy due to its practical advantage of efficiently utilizing already fine-tuned models without requiring joint retraining.
The results indicate that as more fine-tuning tasks are introduced, Speech-FT steadily improves the SUPERB score, demonstrating its ability to absorb and leverage information from different tasks. 

It is also important to note that multitask fine-tuning does not always outperform strong single-task baselines on every evaluation. For example, the PR+SID model yields slightly lower SID accuracy compared to the SID-only model, indicating that negative interference between tasks can indeed occur. Nevertheless, the PR+SID model achieves a higher overall SUPERB score (886.36) than the SID-only model (881.07) as shown in Table~\ref{tab:multitask-compare-single-simplied}. These findings demonstrate that while task-level trade-offs exist, Speech-FT in multitask settings can still enhance overall representation quality.

\begin{table*}[ht]
\centering
\caption{Ablation studies on stable fine-tuning in Speech-FT. Background color highlights cases where the fine-tuning task matches the evaluation task. We use phoneme classification (PC) on TIMIT and speaker identification (SID) on Voxceleb1 as the fine-tuning tasks.}
\scalebox{1.0}{
\begin{tabular}{l l c c c cc c}
\toprule
\multirow{2}{*}{\textbf{Fine-tuning Task}} & \multirow{2}{*}{\textbf{Method}} & \textbf{PR} & \textbf{SID} & \textbf{ER} & \multicolumn{2}{c}{\textbf{SF}} & \multirow{2}{*}{\textbf{$\text{SUPERB}_{\text{S}}\uparrow$}} \\
\cmidrule(r){3-7}
& & \textbf{PER\%$\downarrow$} & \textbf{ACC\%$\uparrow$} & \textbf{ACC\%$\uparrow$} & \textbf{F1$\uparrow$} & \textbf{CER\%$\downarrow$} & \\
\midrule
\multirow{2}{*}{SID on Voxceleb1} & Speech-FT & 6.91 & \cellcolor{gray!20} 88.80 &  66.03 & 87.20 & 25.23 & 881.07 \\
& - Stable fine-tuning & 10.18 &  \cellcolor{gray!20} 88.49 & 63.59 & 86.19 & 28.35 & 836.08 \\
\midrule
\multirow{2}{*}{PC on TIMIT} & Speech-FT & 4.76 & 81.78 & 65.48 & 88.65 & 24.05 & 877.66 \\
& - Stable fine-tuning & 4.78 & 81.77 & 65.17 & 88.60 & 24.89 & 872.12 \\
\midrule
\midrule
\multicolumn{2}{c}{Pre-trained} & 5.17 & 81.86 & 64.99 & 88.54 & 24.70 & 870.20 \\
\bottomrule
\end{tabular}
}
\label{tab:ablation}
\end{table*}

\section{Discussion}
\label{sec:discuss}

\subsection{Generality of Speech-FT across Models and Evaluation Tasks}

To investigate the generality of Speech-FT, 
we first examine Speech-FT across speech representation models with diverse pre-training objectives, and then assess its performance under a comprehensive evaluation setting using the full SUPERB benchmark~\cite{yang2021superb}.
We adopt ASR on TED-LIUM~\cite{rousseau2012ted} as the fine-tuning task throughout this subsection. 

\label{sec:generalizability}
\vspace{1em} 
\subsubsection{Speech-FT is effective across different speech representation models}

We first apply Speech-FT to two widely used speech representation models that have distinct pre-training objectives~\cite{mohamed2022self} from HuBERT: wav2vec 2.0~\cite{baevski2020wav2vec}, which is trained with a contrastive objective; and DeCoAR 2.0~\cite{ling2020decoar}, which is based on a generative approach.
The results are shown in Table~\ref{tab:other-ssl}.
Speech-FT consistently improves SUPERB score for both models.
For instance, on DeCoAR 2.0, it increases the score from 736.77 (pre-trained) to 781.77, while Stable-FT decreases it to 701.60. Similarly, on wav2vec 2.0, Speech-FT raises the score from 837.08 to 864.79, whereas Stable-FT lowers it to 824.82.

We further evaluate Speech-FT on WavLM Base+~\cite{chen2022wavlm}, a model that represents the state of the art among base-sized (Transformers of 12 layers) speech representation models.
Notably, WavLM Base+ is pre-trained on 94,000 hours of unlabelled speech, in contrast to the 960 hours used for HuBERT, wav2vec 2.0 and DeCoAR 2.0, resulting in significantly more general speech representations.
Although the pre-trained model already demonstrates strong performance, Speech-FT is still able to deliver further improvements.
In particular, the overall SUPERB score improves from 942.96 to 945.50.

\vspace{1em} 
\subsubsection{Speech-FT remains effective on full SUPERB evaluation}

We further evaluate Speech-FT on all ten tasks in the SUPERB benchmark, reporting the comprehensive SUPERB\textsubscript{FULL} score. 
Unlike the SUPERB\textsubscript{S} scores used elsewhere in the paper, SUPERB\textsubscript{FULL} encompasses the complete set of tasks of SUPERB when computing Equation~\ref{eq:superb-score}. 
The task definitions and evaluation metrics follow the original SUPERB paper~\cite{yang2021superb}.

We select HuBERT and WavLM Base+ as representative models. The results are presented in Table~\ref{tab:full-superb}. 
Speech-FT significantly enhances the performance of HuBERT, raising its SUPERB\textsubscript{FULL} score from 815.47 to 874.51. 
Consistent improvements are observed across content-related tasks (e.g., ASR: 6.38\% WER to 5.75\%), speaker-related tasks (e.g., ASV: 5.56\% EER to 5.27\%), and semantic-related tasks (e.g., IC: 98.02\% to 98.97\%).

For WavLM Base+, whose pre-trained model already exhibits strong performance, Speech-FT leads to minor drops on speaker-related tasks such as ASV and SD, but still yields gains on most tasks. 
For example, the PER is reduced from 3.94\% to 3.42\% on PR, corresponding to a relative improvement of 13.2\%.
Additionally, the overall SUPERB\textsubscript{FULL} score increases from 946.95 to 950.60. These findings confirm that Speech-FT remains effective under extensive evaluation settings.

\begin{table*}[ht]
\centering
\caption{Comparison of Speech-FT with alternative fine-tuning strategies.
Experiments are conducted on phoneme classification (PC) using the TIMIT dataset and speaker identification (SID) using VoxCeleb1. ``Weight-Space Reg.'' denotes fine-tuning with weight-space regularization. ``LoRA'' indicates LoRA fine-tuning~\cite{hu2022lora}. ``DoRA'' indicates DoRA fine-tuning~\cite{liu2024dora}. ``Early Checkpoint (20\%)'' denotes the model snapshot saved at 20\% of the total fine-tuning steps. The background color highlights cases where the evaluation task is included in the fine-tuning tasks.
}
\scalebox{1.0}{
\begin{tabular}{l l c c c cc c}
\toprule
\multirow{2}{*}{\textbf{Fine-tuning Task}} & \multirow{2}{*}{\textbf{Method}} & \textbf{PR} & \textbf{SID} & \textbf{ER} & \multicolumn{2}{c}{\textbf{SF}} & \multirow{2}{*}{\textbf{$\text{SUPERB}_{\text{S}}\uparrow$}} \\
\cmidrule(r){3-7}
& & \textbf{PER\%$\downarrow$} & \textbf{ACC\%$\uparrow$} & \textbf{ACC\%$\uparrow$} & \textbf{F1$\uparrow$} & \textbf{CER\%$\downarrow$} & \\
\midrule
\multirow{4}{*}{SID on Voxceleb1} & Speech-FT & \textbf{6.91} &  \cellcolor{gray!20} \textbf{88.80} &  \textbf{66.03} & \textbf{87.20} & \textbf{25.23} & \textbf{881.07} \\
& Weight-Space Reg. & 9.94 & \cellcolor{gray!20} 88.34 & 63.88 & 86.13 & 28.24 & 838.56 \\
& LoRA & 13.07 & \cellcolor{gray!20} 87.30 & 62.83 & 83.3 & 33.05 & 785.93 \\
& DoRA & 15.83 & \cellcolor{gray!20} 88.03 & 63.54 & 83.92 & 32.34 & 790.12 \\
& Early Checkpoint (20\%) & 19.75 & \cellcolor{gray!20} 88.33 & 61.80 & 81.04 & 36.39 & 736.06 \\
\midrule
\midrule
\multirow{4}{*}{PC on TIMIT} & Speech-FT & 4.76 & \textbf{81.78} & \textbf{65.48} & \textbf{88.65} & \textbf{24.05} & \textbf{877.66} \\
& Weight-Space Reg. & \textbf{4.46} & 80.67 & 64.84 & 87.91 & 25.41 & 862.26 \\
& LoRA & 4.96 & 71.43 & 63.34 & 88.03 & 25.9 & 824.72 \\
& DoRA & 7.23 & 77.54 & 63.78 & 87.23 & 26.30 & 830.92 \\
 & Early Checkpoint (20\%) & 6.33 & 71.80 & 61.58 & 86.43 & 28.63 & 790.44 \\
\midrule
\midrule
\multicolumn{2}{c}{Pre-trained} & 5.17 & 81.86 & 64.99 & 88.54 & 24.70 & 870.20 \\
\bottomrule
\end{tabular}
}
\label{tab:regularization-full}
\end{table*}
\subsection{Stable-FT Enhances Cross-Task Generalization in Speech-FT}
\label{sec:ablation}
To understand the impact of Stable-FT in Speech-FT, we conduct an ablation study by removing it from the Speech-FT framework.
This analysis is conducted on phoneme classification (PC) on TIMIT and speaker identification (SID) on VoxCeleb1.
As shown in Table~\ref{tab:ablation}, removing Stable-FT significantly impairs the cross-task generalization ability of the model.

Specifically, when fine-tuning with SID, removing Stable-FT leads to notable performance degradation on unrelated tasks: PR increases by 3.27\% PER, ER drops by 2.44\% ACC, and SF increases by 3.12\% CER. In this setting, the model even fails to outperform the pre-trained baseline in terms of the SUPERB score.

When fine-tuning with PC, the impact is less pronounced but remains evident: ER drops by 0.31\% ACC, and SF increases by 0.84\% CER. Although smaller in magnitude, these changes still negatively affect the model’s cross-task generalization ability, leading to a decrease in the SUPERB score.
These results underscore the essential role of Stable-FT in enabling Speech-FT to preserve cross-task generalization.

\subsection{Comparison to Baselines That Constrain Weight Deviation During Fine-tuning} 
\label{sec:strong-baselines}

In this section, we compare Speech-FT with several fine-tuning baselines that aim to constrain weight deviations during fine-tuning.
These methods may reduce deviations in weight space; however, they do not necessarily maintain feature similarity with the pre-trained model. As a result, they may lose cross-task generalization ability after fine-tuning.
Specifically, we apply several possible alternatives of weight-space interpolation \emph{on top of Stable-FT}, including weight-space regularization, LoRA fine-tuning, and early checkpoints during the fine-tuning process.

All methods are applied to two representative fine-tuning tasks: TIMIT for phoneme classification (PC) and VoxCeleb1 for speaker identification (SID).
Our goal is to determine whether these alternatives can match Speech-FT in preserving cross-task generalization ability.
A detailed analysis of the underlying differences between these alternatives and Speech-FT is provided in Section~\ref{sec:property-analysis}.

\vspace{1em} 
\subsubsection{Speech-FT outperforms weight-space regularization}
Weight-space regularization offers a straightforward alternative to our interpolation-based approach by explicitly constraining parameter updates. 
During fine-tuning, we add a regularization term $\lambda\|\theta' - \theta_0\|^2_F$ to penalize deviation from the pre-trained weights. The hyperparameter $\lambda$ is selected from $\{10^{-2}, 10^{-4}, 10^{-6}\}$, and we report the best-performing configuration.

As shown in Table~\ref{tab:regularization-full}, weight-space regularization improves performance on tasks aligned with the fine-tuning objective, but consistently degrades performance on those that are not.  
For instance, when fine-tuned on PC, models trained with regularization perform worse than both Speech-FT and the pre-trained model on unrelated tasks such as SID and ER. SID accuracy drops to 80.67\% and ER accuracy to 64.84\%, both lower than the pre-trained scores of 81.86\% and 64.99\%, respectively.  
Similarly, when fine-tuned on SID, PER increases to 9.94\% on PR, and ER accuracy drops to 63.88\%, again underperforming both the pre-trained model and Speech-FT.

As a result, the overall SUPERB scores of regularized models are not only lower than those of Speech-FT, but also fall short of the pre-trained baseline.  
When fine-tuned on PC, weight-space regularization yields a SUPERB score of 862.26, compared to 877.66 from Speech-FT and 870.20 from the pre-trained model.  
When fine-tuned on SID, the regularized model scores 838.56—substantially below both Speech-FT at 881.07 and the pre-trained model at 870.20.
These results suggest that weight-space regularization is insufficient to preserve the cross-task generalization ability of the pre-trained representations.

\begin{table}[ht]
\centering
\caption{The impact of hyperparameter $\alpha$ when fine-tuning on phoneme classification (TIMIT), speaker identification (VoxCeleb1), and emotion recognition (IEMOCAP).}
\scalebox{1.0}{
\begin{tabular}{l ccc}
\toprule
\multirow{2}{*}{\textbf{$\alpha$}} & \textbf{PC on TIMIT} & \textbf{SID on Voxceleb1} & \textbf{ER on IEMOCAP} \\
\cmidrule{2-4}
& \textbf{SUPERB$_s$}$\uparrow$ & \textbf{SUPERB$_s$}$\uparrow$ & \textbf{SUPERB$_s$}$\uparrow$ \\
\midrule
0.0  & 870.20 & 870.20 & 870.2 \\
0.10 & 874.68 & 876.89 & 932.59 \\
0.25 & \textbf{877.66} & \textbf{881.07} & \textbf{1010.29} \\
0.50 & 843.75 & 810.36 & 1007.38 \\
0.75 & 810.79 & 721.54 & 983.72 \\
1.0  & 726.64 & 651.42 & 908.86 \\
\bottomrule
\end{tabular}
}
\label{tab:hyperparameter}
\end{table}

\vspace{1em}
\subsubsection{Speech-FT outperforms parameter-efficient fine-tuning methods}
To further contextualize the effectiveness of Speech-FT, we compare it with LoRA fine-tuning~\cite{hu2022lora} as an additional baseline.
LoRA was originally proposed to improve fine-tuning efficiency by introducing low-rank adaptation modules while keeping the pre-trained weights fixed. 
It also serves as a relevant baseline for Speech-FT, as it similarly aims to minimize changes to the pre-trained model. 
However, while LoRA constrains changes in weight space, Speech-FT instead constrains changes in feature space.
We follow the implementation of prior work~\cite{chen2023explore} for LoRA fine-tuning on speech representation models.

As shown in Table~\ref{tab:regularization-full}, LoRA consistently underperforms Speech-FT across all tasks, with particularly large performance drops on unrelated evaluation tasks. 
For instance, when fine-tuned on PC, LoRA achieves 71.43\% accuracy on SID, 63.34\% on ER, and a SUPERB score of 824.72. In comparison, Speech-FT achieves 81.78\%, 65.48\%, and 877.66, respectively.

A similar pattern is observed when fine-tuning on SID. In this setting, LoRA yields 9.94\% PER on PR, 63.88\% accuracy on ER, and a SUPERB score of 838.56.
In contrast, Speech-FT achieves 6.91\% PER, 66.03\% accuracy, and a significantly higher SUPERB score of 881.07.
These results indicate that although LoRA introduces minimal additional parameters, it still causes degraded cross-task generalization. 

We also experiment with DoRA fine-tuning~\cite{liu2024dora}, a recently proposed PEFT method that improves on LoRA by decomposing weights into magnitude and direction. While DoRA achieves slightly higher SUPERB scores than LoRA, it nevertheless suffers from the same drawback and shows severely degraded cross-task generalization compared to Speech-FT.

\vspace{1em}
\subsubsection{Speech-FT outperforms early checkpoints during fine-tuning}
We also examine whether stopping the fine-tuning process early could help preserve cross-task generalization. 
Specifically, we compare Speech-FT with checkpoints obtained at 20\%, 40\%, and 60\% of the total fine-tuning steps. 
These checkpoints represent intermediate stages of fine-tuning, where the model has not yet fully adapted its parameters and thus remains relatively close to the pre-trained model in weight space.  

Among the 20\%, 40\%, and 60\% checkpoints, the 20\% checkpoint yields the most favorable results, which we report in Table~\ref{tab:regularization-full}.
Although early checkpoints could introduce fewer deviations in weight space compared to full fine-tuning, they still lead to degraded performance across all evaluation tasks.
When fine-tuned on PC, the 20\% checkpoint yields a SUPERB score of 790.44, while SID accuracy drops from 81.86\% (pre-trained model) to 71.80\%.
Similarly, when fine-tuned on SID, the SUPERB score decreases to 736.06, accompanied by an increase in PER on PR from 5.17\% (pre-trained model) to 19.75\%.
These findings suggest that early stopping alone is insufficient to preserve cross-task generalization.

\begin{figure}[!t]
    \centering
    \includegraphics[width=0.8\linewidth]{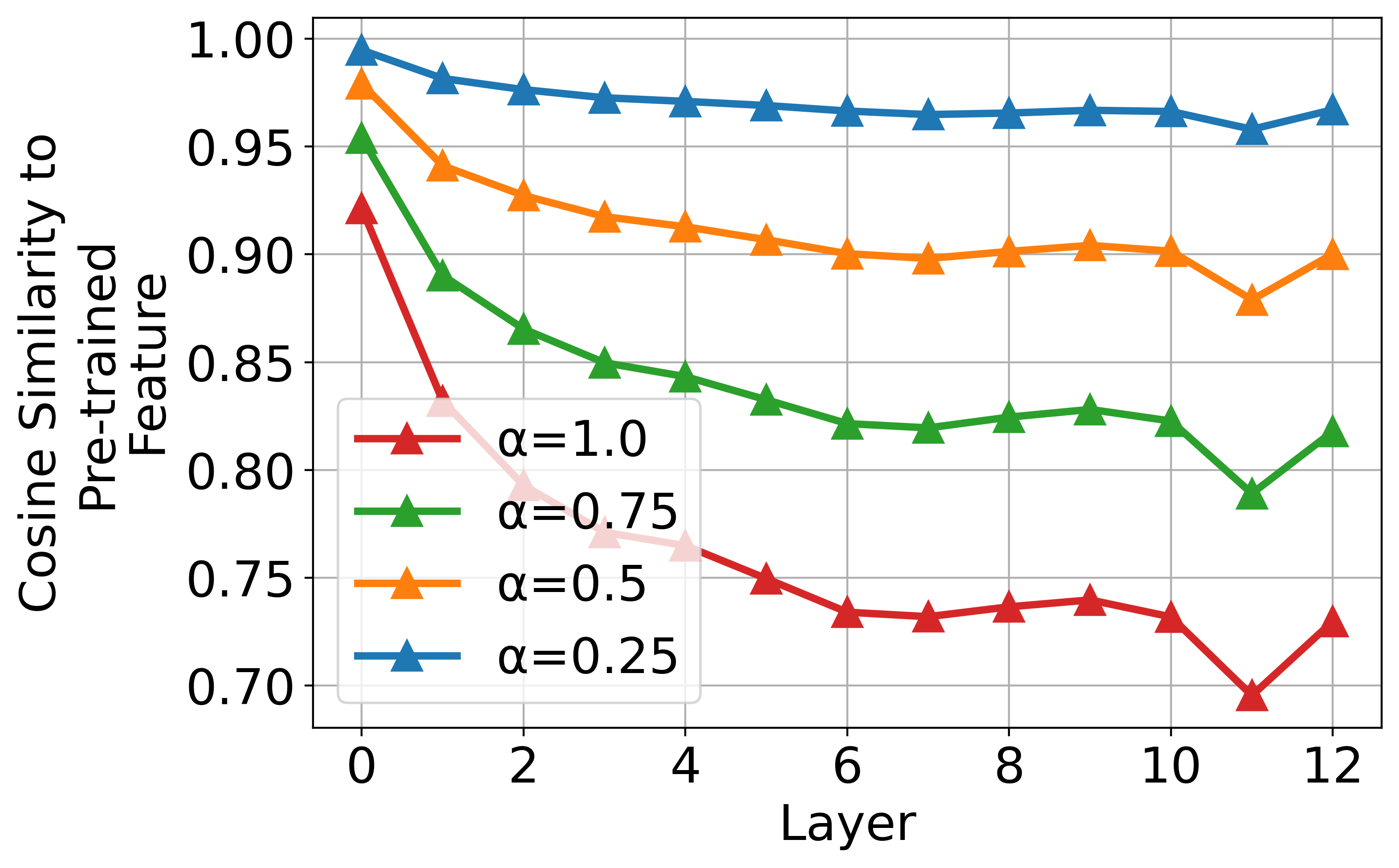}
    \includegraphics[width=0.8\linewidth]{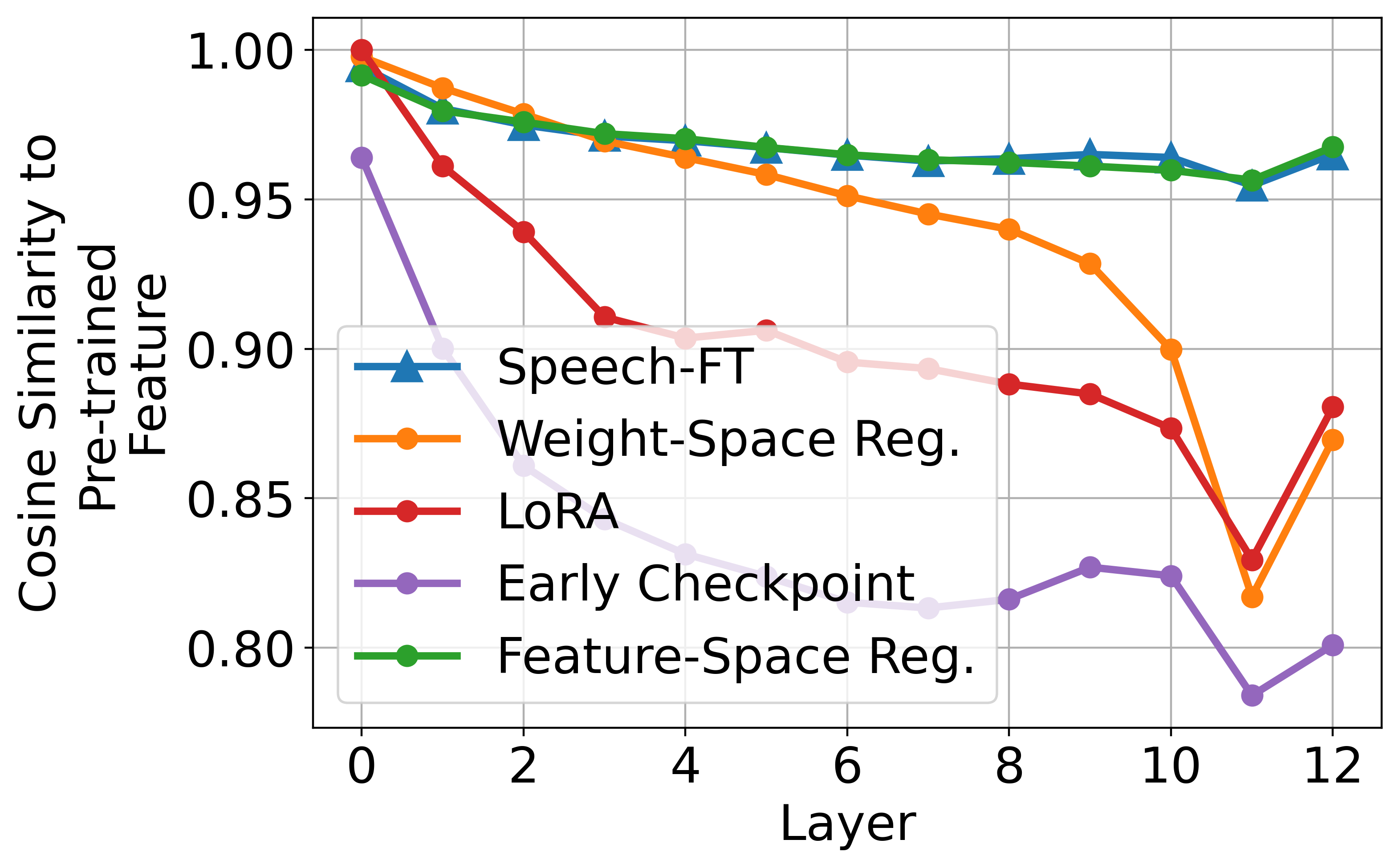}
    \caption{Feature similarity with the pre-trained model. (Top) Effect of $\alpha$ on the cosine similarity between Speech-FT and pre-trained features. (Bottom) Cosine similarity between the pre-trained features and those from Speech-FT, weight-space regularization (``Weight-Space Reg.''), LoRA fine-tuning (``LoRA''), early checkpoint during fine-tuning (``Early Checkpoint''), and feature-space regularization (``Feature-Space Reg.'').}
    \label{fig:feature-sim}
\end{figure}
\begin{figure}[!t]
    \centering
    \includegraphics[width=0.8\linewidth]{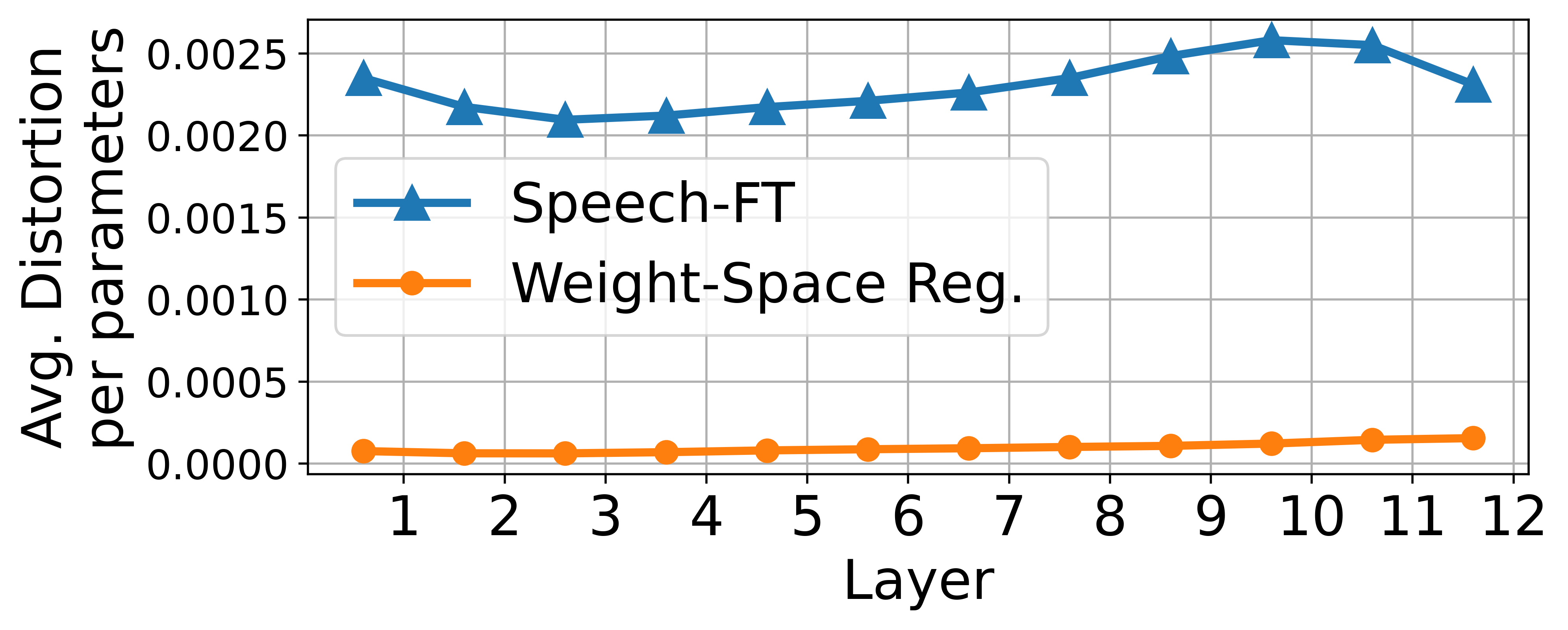}
    \caption{Average L2 distortion per parameter in the weight space with respect to the pre-trained model. ``Weight-Space Reg.'' denotes weight-space regularization.}
    \label{fig:tv-norm}
\end{figure}

\subsection{Speech-FT Preserves Feature Similarity Better Than Alternatives}
\label{sec:property-analysis}

In the previous section, we showed that Speech-FT achieves better cross-task generalization than other alternatives that aim to minimize deviations in weight space. 
To better understand why Speech-FT leads to improved cross-task generalization, we analyze the feature similarity between Speech-FT and the pre-trained model in this section. 
Motivated by the findings, we further introduce a feature-space regularization baseline to investigate whether constraining representational drift via a regularization term during fine-tuning can help preserve cross-task generalization. 
Finally, we provide a brief theoretical explanation of Speech-FT.
All analyses are conducted on the TIMIT dataset for phoneme classification (PC).

\vspace{1em}
\subsubsection{Weight-space interpolation effectively restores feature similarity to the pre-trained model}
Table~\ref{tab:hyperparameter} shows how the performance of speech representations evolves as the interpolation hyperparameter $\alpha$ decreases from 1.0 (fully fine-tuned model) to 0.0 (pre-trained model).
Fine-tuning without weight-space interpolation ($\alpha=1.0$) compromises the cross-task generalization ability of the speech representation. 
However, as $\alpha$ decreases, the SUPERB score gradually improves, reaching its peak at $\alpha = 0.25$.
As shown in the top panel of Figure~\ref{fig:feature-sim}, smaller values of $\alpha$ also lead to higher feature similarity with the pre-trained model. 
This helps explain the observed improvement in cross-task generalization when using lower values of $\alpha$.

\begin{table*}[ht]
\centering
\caption{Comparison of Speech-FT and feature-space regularization (``Feature-Space Reg.''). We use phoneme classification on TIMIT as the fine-tuning task for this experiment.}
\scalebox{1.0}{
\begin{tabular}{l c c c cc c}
\toprule
\multirow{2}{*}{\textbf{Method}} & \textbf{PR} & \textbf{SID} & \textbf{ER} & \multicolumn{2}{c}{\textbf{SF}} & \multirow{2}{*}{\textbf{$\text{SUPERB}_{\text{S}}\uparrow$}} \\
\cmidrule(r){2-6}
& \textbf{PER\%$\downarrow$} & \textbf{ACC\%$\uparrow$} & \textbf{ACC\%$\uparrow$} & \textbf{F1$\uparrow$} & \textbf{CER\%$\downarrow$} & \\
\midrule
Speech-FT & \textbf{4.76} & 81.78 & \textbf{65.48} & \textbf{88.65} & \textbf{24.05} & \textbf{877.66} \\
Feature-Space Reg. & 5.53 & 81.72 & 65.01 & 87.76 & 26.44 & 858.39 \\
\midrule
\midrule
Pre-trained & 5.17 & \textbf{81.86} & 64.99 & 88.54 & 24.70 & 870.20 \\
\bottomrule
\end{tabular}
}

\label{tab:feat-regularization}
\end{table*}

\vspace{1em}
\subsubsection{Speech-FT exhibits higher feature similarity to the pre-trained model than other alternatives}

We further compare how similar the features of Speech-FT and other alternatives are to those of the pre-trained model.
Although Speech-FT results in a larger change in weight space compared to weight-space regularization—as shown in Figure~\ref{fig:tv-norm}—it achieves much higher similarity at the feature level\footnote{In addition to cosine similarity, we also measured the L2 distance between features, and the overall trend remains consistent.} as shown in the bottom panel of Figure~\ref{fig:feature-sim}. 
It is worth noting that the regularization strength for weight-space regularization is a tunable hyperparameter, and all analyses in Figures~\ref{fig:feature-sim} and~\ref{fig:tv-norm} are based on its best-performing configuration.

These findings highlight a key difference between Speech-FT and weight-space regularization. While weight-space regularization could keep the fine-tuned model close to the pre-trained model in parameter space, it fails to preserve feature similarity.
In contrast, Speech-FT, despite allowing more parameter changes, is better at maintaining the original feature space.
This helps explain the results in Table~\ref{tab:regularization-full}, where weight-space regularization degrades cross-task generalization more severely than Speech-FT.
The reduced feature similarity in regularized models likely contributes to their weaker performance on tasks unrelated to the fine-tuning objective.
Moreover, our layer-wise analysis in Figure~\ref{fig:feature-sim} shows that Speech-FT consistently preserves higher similarity in the middle and upper layers, which are known to encode richer phonetic information~\cite{pasad2021layer}. This pattern suggests that Speech-FT may better retain phonetic knowledge from pre-training, consistent with its strong performance on the PR task in Table~\ref{tab:regularization-full}.

In addition, we compare Speech-FT with the other two alternative strategies: LoRA fine-tuning and early checkpoints during fine-tuning.
Both approaches aim to limit parameter modifications to the pre-trained model during fine-tuning.
However, as shown in Figure~\ref{fig:feature-sim}, both methods result in substantially lower feature similarity to the pre-trained model compared to Speech-FT.
This finding, consistent with our earlier observations on weight-space regularization, suggests that merely constraining weight updates is insufficient to maintain the original feature space.

Overall, weight-space interpolation in Speech-FT proves significantly more effective at preserving feature similarity with the pre-trained model than other alternatives.

\vspace{1em}
\subsubsection{Speech-FT outperforms feature-space regularization}

Motivated by the observation in Figure~\ref{fig:feature-sim}, we further compare Speech-FT with a feature-space regularization baseline. This baseline introduces an additional loss term during fine-tuning that penalizes deviations in the model’s hidden representations from those of the pre-trained model.
Specifically, let $f'$ denote the feature produced by the model during fine-tuning and $f_0$ denote the corresponding feature from the pre-trained model.
The regularization term takes the form $\lambda \|f' - f_0\|^2_F$. In practice, this regularization is applied to the hidden features across all layers. Same as weight-space regularization, the hyperparameter $\lambda$ is tuned among $\{10^{-2}, 10^{-4}, 10^{-6}\}$, and we report the best-performing setting. 

As shown in the bottom panel of Figure~\ref{fig:feature-sim}, feature-space regularization successfully maintains a high degree of similarity to the pre-trained model across all layers, as measured by both cosine similarity and L2 distance. This suggests that the feature-space regularization objective is effective in constraining the representations to remain close to those learned during pre-training.

Correspondingly, Table~\ref{tab:feat-regularization} shows that this method could help preserve cross-task generalization ability. 
Specifically, feature-space regularization achieves 81.72\% accuracy on SID and 65.01\% accuracy on ER, which are comparable to the performance of the pre-trained model. Compared to the results of weight-space regularization reported in Table~\ref{tab:regularization-full}, which achieve 80.67\% on SID and 64.84\% on ER, feature-space regularization yields notable improvements on tasks unrelated to the fine-tuning objective.

However, feature-space regularization fails to yield improvements on the related task.
For example, it results in a PER of 5.53\% on PR, compared to 4.76\% achieved by Speech-FT.
This suggests that while adding a regularization term to constrain representational drift may help preserve cross-task generalization, it can also hinder the model’s ability to learn task-specific information.
In contrast, Speech-FT does not impose any explicit regularization during fine-tuning; instead, it restores feature similarity with the pre-trained model through subsequent interpolation, thereby achieving a markedly better trade-off between the learning of task-relevant information and the preservation of cross-task generalization ability.

\vspace{1em}
\subsubsection{Theoretical explanation of Speech-FT}
\begin{table}[ht]
\centering
\caption{LLFC verification under Speech-FT. Endpoints ($\alpha\in\{0,1\}$) omitted for brevity. Features are orthogonally aligned to the pre-trained endpoint. Higher $R^2$ indicates stronger layerwise linearity, and regression coefficients $(b_0,b_1)$ closer to the theoretical $(1-\alpha,\alpha)$ indicate better agreement with the predicted linear mixture.}
\scalebox{1.0}{
\begin{tabular}{l c c c c}
\toprule
\multirow{2}{*}{\textbf{Method}} & \multirow{2}{*}{$\boldsymbol{\alpha}$} & \multicolumn{3}{c}{\textbf{Linear mixture fit}} \\
\cmidrule(lr){3-5}
& & $\mathbf{b_0}$ & $\mathbf{b_1}$ & $\mathbf{R^2}$$\uparrow$ \\
\midrule
\multirow{3}{*}{Speech-FT}
& 0.25 & 0.782 & 0.220 & 0.985  \\
& 0.50 & 0.491 & 0.507 & 0.980  \\
& 0.75 & 0.219 & 0.779 & 0.989  \\
\midrule
\multirow{3}{*}{\makecell[l]{Speech-FT\\- Stable fine-tuning}}
& 0.25 & 0.831 & 0.185 & 0.985  \\
& 0.50 & 0.536 & 0.474 & 0.977  \\
& 0.75 & 0.239 & 0.763 & 0.986  \\
\bottomrule
\end{tabular}
}
\label{tab:llfc}
\end{table}

The theoretical basis of Speech-FT can be explained through the lens of Layerwise Linear Feature Connectivity (LLFC)~\cite{zhou2023going}. 
LLFC was originally defined for image classification models fine-tuned from the same pre-trained initialization, where internal representations evolve almost linearly along the interpolation path.
In our setting, one endpoint is the pre-trained model $\theta_0$ and the other is its fine-tuned variant $\theta'$. Since $\theta'$ is directly obtained from $\theta_0$, the same LLFC principle applies: the per-layer features along the interpolation path $F_\ell(\hat{\theta}(\alpha)) \approx (1-\alpha)F_\ell(\theta_{0}) + \alpha F_\ell(\theta')$ are well approximated by the linear mixture of the two endpoints’ features, indicating that representation geometry is largely preserved.

To verify this in practice, we evaluate LLFC on a development set by extracting per-layer features $F_\ell(\cdot)$ from $\theta_0$, $\theta'$, and interpolated checkpoints $\hat{\theta}(\alpha)$ ($\alpha \in \{0.25,0.50,0.75\}$). 
After aligning these features to $F_\ell(\theta_{0})$ via Orthogonal Procrustes~\cite{schonemann1966procrustes}, thereby removing arbitrary rotations between representation spaces, we compute a ridge regression~\cite{hoerl1970ridge} of $F_\ell(\hat{\theta}(\alpha))$ on $\{F_\ell(\theta_0),F_\ell(\theta')\}$, yielding coefficients $(b_0,b_1)$ and the coefficient of determination $R^2$, with higher values indicating a better fit. 
As summarized in Table~\ref{tab:llfc}, Speech-FT exhibits high $R^2$ values and regression coefficients close to the theoretical $(1-\alpha,\alpha)$, confirming that it satisfies LLFC. Additionally, the stable fine-tuning step appears to play a crucial role in supporting this behavior.

This means interpolation explicitly pulls the features back toward pre-training directions, thereby restoring similarity to the original representations. In contrast, weight-space regularization only constrains parameter distance without ensuring that the resulting models lie in a representation-preserving region. As a result, it does not guarantee recovery of the pre-training feature geometry.

While LLFC explains the mechanism for restoring feature similarity, the framework of Task Arithmetic~\cite{ilharco2022editing} provides another perspective for why task-specific knowledge is simultaneously retained. This framework treats the changes from fine-tuning as a \textit{task vector}—the arithmetic difference between the fine-tuned and pre-trained model weights—which encapsulates the knowledge learned for a specific task. By defining this task vector as $\tau=\theta'-\theta_0$, our interpolation is equivalent to: $\hat{\theta}(\alpha)=\theta_0+\alpha\tau$. This clarifies that Speech-FT operates by adding a scaled version of the task-specific knowledge ($\tau$) to the pre-trained base model ($\theta_0$). This process inherently preserves the general capabilities of $\theta_0$ while controllably injecting the specialized knowledge contained within $\tau$.

\section{Related Work}

\subsection{Model Merging}
Model merging is a technique that combines multiple models into a single model while preserving the key attributes of each original model~\cite{ilharco2022editing}. Weight-space interpolation can be viewed as a form of model merging, where a pre-trained and a fine-tuned model are combined in parameter space.

Numerous studies have explored model merging in computer vision, including techniques for multi-target domain adaptation~\cite{li2024training} and maintaining task-specific information through selective parameter merging~\cite{marczak2024magmax}. In natural language processing, model merging methods have been applied to mitigate the alignment tax of RLHF via model averaging~\cite{lin2024mitigating} and enhance multi-task learning with large language models~\cite{zhou2024metagpt}.

In the speech domain, several studies have investigated the applications of model merging. These include merging ASR models trained on different dataset distributions to enhance out-of-distribution generalization~\cite{ramesh2024task, plantinga2024parameter}, merging TTS models to interpolate speaker attributes~\cite{murata2024attribute}, merging speech translation models to expand language pairs~\cite{cheng2024task}, and merging fine-tuned ASR models to support lifelong learning of unseen accents~\cite{kulshreshtha2024sequential}.

In contrast to these prior works, we propose a novel application of model merging for speech representation learning—an unexplored scenario. Our approach not only introduces model merging into this domain but also provides insights into how model merging can maintain cross-task generalization ability, offering a fresh perspective on its benefits.

\subsection{Out-of-Distribution Generalization}
Out-of-distribution (OOD) generalization aims to preserve a model's performance on OOD datasets after fine-tuning.
Several studies in computer vision have explored OOD generalization for vision pre-trained models that could perform zero-shot inference.
For instance, LP-FT~\cite{kumar2022fine} introduces a simple two-step approach—linear probing followed by full fine-tuning—to mitigate distortion in pre-trained features by first learning a near-optimal linear head during the initial steps.
Model Soup~\cite{wortsman2022model} demonstrates that averaging the weights of models fine-tuned with different hyperparameter configurations can improve robustness on OOD datasets.

A relevant approach to Speech-FT in a different context is Wise-FT~\cite{wortsman2022robust}, which also employs weight-space interpolation between pre-trained and fine-tuned models. However, it differs in several key aspects.
First, Wise-FT is specifically designed to enhance out-of-distribution (OOD) generalization in image classification, where the model is tested under distributional shifts such as changes in style or image perturbations. In contrast, Speech-FT addresses a more complex challenge—preserving generalization across diverse downstream tasks in speech representation learning. Second, Speech-FT introduces a Stable-FT stage before interpolation, which is shown in Section~\ref{sec:ablation} to be critical for maintaining cross-task generalization. Notably, using weight-space interpolation alone may be insufficient to improve speech representation. 
Third, we evaluate the effectiveness of Speech-FT across diverse fine-tuning tasks in speech processing.
We also evaluate Speech-FT under unsupervised language adaptation and in scenarios involving multiple fine-tuning tasks. 
These explorations have not been addressed in prior works.
Finally, our study provides an analysis of why weight-space interpolation is effective in speech representation learning, offering insights that are absent from prior works.

\subsection{Continual Learning}
Continual learning, or lifelong learning, enables models to adapt to new domain data while retaining previously acquired knowledge. It addresses the issue of catastrophic forgetting~\cite{kirkpatrick2017overcoming} and is essential for handling dynamic and evolving environments. Continual learning typically involves multiple rounds of fine-tuning to simulate real-world scenarios where models continually learn from new data.

Several prior studies have explored continual learning for automatic speech recognition~\cite{kulshreshtha2024sequential, houston2020continual}.
While we also aim to preserve previously acquired information, our objective is fundamentally different from that of continual learning. Rather than simulating a long sequence of tasks, we focus primarily on enhancing speech representations after a single fine-tuning stage. Since our main goal is to improve representation quality—rather than to ensure robustness under sequential task adaptation—it is unnecessary to introduce task sequences into our setting.

\subsection{Fine-tuning Speech Representation Models}

Several prior works have attempted to fine-tune speech representation models, with most focusing on improving performance on specific downstream tasks, such as speech recognition~\cite{yang2024finetuning, chen2023explore}, emotion recognition~\cite{chen2023emotionfinetune}, and anti-spoofing detection~\cite{pan24c_interspeech, guragain2024speech}.
In contrast, our work falls within the domain of speech representation learning—we aim to improve the overall quality of speech representations through fine-tuning, with a focus on enhancing performance across a broad range of speech-related tasks.

One prior study explored improving speech representations via supervised fine-tuning tasks~\cite{chen2021speech}, but it did not explicitly address the preservation of cross-task generalization. As a result, their approach fails to generalize well across tasks.

Another study investigated continuing the pre-training process on a low-resource language~\cite{getman2024happens}, but their focus was on improving performance in that specific language. In contrast, our goal is to preserve generalization across both previously learned and new languages. 

\section{Conclusion}  

We have presented Speech-FT, a simple yet effective fine-tuning strategy that enhances speech representation models while preserving their overall cross-task generalization ability.
Speech-FT improves representations across diverse fine-tuning scenarios and model types, without introducing any additional computational cost compared to standard fine-tuning, making it both practical and efficient.
Our extensive experiments demonstrate that Speech-FT consistently outperforms other fine-tuning baselines, including regularization-based approaches and parameter-efficient fine-tuning techniques.
Looking ahead, we believe that Speech-FT may offer broader applicability in other domains, such as spoken language modeling and beyond.

\section*{Acknowledgements}
Tzu-Quan Lin is supported by a joint Ph.D. scholarship from NTU GICE and Delta Electronics, Inc.

\bibliographystyle{IEEEtran}
\bibliography{custom}

@inproceedings{kulshreshtha2024sequential,
  title     = {{Sequential Editing for Lifelong Training of Speech Recognition Models}},
  author    = {Kulshreshtha, Devang and Pappas, Nikolaos and Houston, Brady and Dingliwal, Saket and Ronanki, Srikanth},
  year      = {2024},
  booktitle = {Interspeech 2024},
  pages     = {3919--3923},
  doi       = {10.21437/Interspeech.2024-2027},
  issn      = {2958-1796},
}

@inproceedings{houston2020continual,
  title     = {{Continual Learning for Multi-Dialect Acoustic Models}},
  author    = {Houston, Brady and Kirchhoff, Katrin Kirchhoff},
  year      = {2020},
  booktitle = {Interspeech 2020},
  pages     = {576--580},
  doi       = {10.21437/Interspeech.2020-1797},
  issn      = {2958-1796},
}

@inproceedings{baevski2020wav2vec,
 author = {Baevski, Alexei and Zhou, Yuhao and Mohamed, Abdelrahman and Auli, Michael},
 booktitle = {The Thirty-Fourth Annual Conference on Neural Information Processing Systems},
 pages = {12449--12460},
 title = {{wav2vec 2.0: A Framework for Self-Supervised Learning of Speech Representations}},
 volume = {33},
 year = {2020}
}

@article{hsu2021hubert,
  author={Hsu, Wei-Ning and Bolte, Benjamin and Tsai, Yao-Hung Hubert and Lakhotia, Kushal and Salakhutdinov, Ruslan and Mohamed, Abdelrahman},
  journal={IEEE/ACM Transactions on Audio, Speech, and Language Processing}, 
  title={{HuBERT: Self-Supervised Speech Representation Learning by Masked Prediction of Hidden Units}}, 
  year={2021},
  volume={29},
  pages={3451-3460}
}

@article{chen2022wavlm,
  author={Chen, Sanyuan and others},
  journal={IEEE Journal of Selected Topics in Signal Processing}, 
  title={{WavLM: Large-Scale Self-Supervised Pre-Training for Full Stack Speech Processing}}, 
  year={2022},
  volume={16},
  number={6},
  pages={1505-1518}
}

@article{ling2020decoar,
  author    = {Ling, Shaoshi and Liu, Yuzong},
  title     = {{DeCoAR 2.0: Deep Contextualized Acoustic Representations with Vector Quantization}},
  journal   = {arXiv preprint arXiv:2012.06659},
  year      = {2020}
}

@article{kirkpatrick2017overcoming,
  author  = {Kirkpatrick, James and others},
  title   = {{Overcoming Catastrophic Forgetting in Neural Networks}},
  journal = {Proceedings of the National Academy of Sciences},
  volume  = {114},
  number  = {13},
  pages   = {3521--3526},
  year    = {2017},
  doi     = {10.1073/pnas.1611835114}
}

@article{chen2021speech,
  author  = {Chen, Yi-Chen and Yang, Shu-wen and Lee, Cheng-Kuang and See, Simon and Lee, Hung-yi},
  title   = {{Speech Representation Learning Through Self-Supervised Pretraining and Multi-Task Finetuning}},
  journal = {arXiv preprint arXiv:2110.09930},
  year    = {2021},
  doi     = {10.48550/arXiv.2110.09930}
}

@inproceedings{getman2024happens,
  author    = {Getman, Yaroslav and Gr{\'o}sz, Tam{\'a}s and Kurimo, Mikko},
  title     = {{What Happens in Continued Pre-Training? Analysis of Self-Supervised Speech Models with Continued Pre-Training for Colloquial Finnish ASR}},
  booktitle = {Interspeech 2024},
  pages     = {5043--5047},
  year      = {2024},
  doi       = {10.21437/Interspeech.2024-476},
  issn      = {2958-1796}
}

@inproceedings{yang2021superb,
  author    = {Yang, Shu-wen and others},
  title     = {{SUPERB: Speech Processing Universal PERformance Benchmark}},
  booktitle = {Interspeech 2021},
  pages     = {1194--1198},
  year      = {2021},
  doi       = {10.21437/Interspeech.2021-1775},
  issn      = {2958-1796}
}

@inproceedings{feng2023superb,
  author    = {Feng, Tzu-hsun and others},
  title     = {{SUPERB@SLT 2022: Challenge on Generalization and Efficiency of Self-Supervised Speech Representation Learning}},
  booktitle = {2022 IEEE Spoken Language Technology Workshop (SLT)},
  pages     = {1096--1103},
  year      = {2023},
  publisher = {IEEE},
  doi       = {10.1109/SLT54892.2023.10067243}
}

@article{mohamed2022self,
  author={Mohamed, Abdelrahman and others},
  journal={IEEE Journal of Selected Topics in Signal Processing}, 
  title={{Self-Supervised Speech Representation Learning: A Review}}, 
  year={2022},
  volume={16},
  number={6},
  pages={1179-1210},
  doi={10.1109/JSTSP.2022.3207050}
}

@inproceedings{wortsman2022robust,
  author    = {Wortsman, Mitchell and others},
  title     = {{Robust Fine-Tuning of Zero-Shot Models}},
  booktitle = {Proceedings of the IEEE/CVF Conference on Computer Vision and Pattern Recognition (CVPR)},
  pages     = {7959--7971},
  year      = {2022},
  doi       = {10.1109/CVPR52688.2022.00781}
}

@inproceedings{kumar2022fine,
  author  = {Kumar, Ananya and Raghunathan, Aditi and Jones, Robbie and Ma, Tengyu and Liang, Percy},
  title   = {{Fine-Tuning Can Distort Pretrained Features and Underperform Out-of-Distribution}},
  pages = {1--15},
  booktitle={The Tenth International Conference on Learning Representations},
  year={2022}
}

@inproceedings{wortsman2022model,
  author    = {Wortsman, Mitchell and others},
  title     = {{Model Soups: Averaging Weights of Multiple Fine-Tuned Models Improves Accuracy Without Increasing Inference Time}},
  booktitle = {Proceedings of the 39th International Conference on Machine Learning},
  pages     = {23965--23998},
  year      = {2022},
  publisher = {PMLR},
  volume    = {162},
}

@inproceedings{lin2024mitigating,
  author    = {Lin, Yong and others},
  title     = {{Mitigating the Alignment Tax of RLHF}},
  booktitle = {Proceedings of the 2024 Conference on Empirical Methods in Natural Language Processing (EMNLP)},
  pages     = {580--606},
  year      = {2024},
  doi       = {10.18653/v1/2024.emnlp-main.35}
}

@inproceedings{ilharco2022editing,
  author    = {Ilharco, Gabriel and others},
  title     = {{Editing Models with Task Arithmetic}},
  booktitle = {The Eleventh International Conference on Learning Representations},
  year      = {2023},
  pages = {1--17}
}

@inproceedings{lin2023melhubert,
  author    = {Lin, Tzu-Quan and Lee, Hung-yi and Tang, Hao},
  title     = {{MelHuBERT: A Simplified HuBERT on Mel Spectrograms}},
  booktitle = {2023 IEEE Automatic Speech Recognition and Understanding Workshop (ASRU)},
  year      = {2023},
  pages     = {1-8},
  publisher = {IEEE},
  doi       = {10.1109/ASRU57964.2023.10389700}
}

@inproceedings{wang2023minisuperb,
  author    = {Wang, Yu-Hsiang and Chen, Huang-Yu and Chang, Kai-Wei and Hsu, Winston and Lee, Hung-yi},
  title     = {{MiniSUPERB: Lightweight Benchmark for Self-Supervised Speech Models}},
  booktitle = {2023 IEEE Automatic Speech Recognition and Understanding Workshop (ASRU)},
  year      = {2023},
  pages     = {1-8},
  publisher = {IEEE},
  doi       = {10.1109/ASRU57964.2023.10389699}
}

@article{garofolo1993darpa,
  author  = {Garofolo, John S. and Lamel, Lori F. and Fisher, William M. and Fiscus, Jonathan G. and Pallett, David S. and Dahlgren, Nancy L.},
  title   = {{DARPA TIMIT: Acoustic-Phonetic Continuous Speech Corpus CD-ROM. NIST Speech Disc 1-1.1}},
  journal = {NASA STI/Recon Tech. Rep. N},
  year    = {1993},
  volume  = {93},
  note    = {{Art.} no. 27403}
}

@inproceedings{panayotov2015librispeech,
  author    = {Panayotov, Vassil and Chen, Guoguo and Povey, Daniel and Khudanpur, Sanjeev},
  title     = {{LibriSpeech: An ASR Corpus Based on Public Domain Audio Books}},
  booktitle = {ICASSP 2015 - 2015 IEEE International Conference on Acoustics, Speech and Signal Processing (ICASSP)},
  pages     = {5206--5210},
  year      = {2015},
  publisher = {IEEE},
  doi       = {10.1109/ICASSP.2015.7178964}
}

@article{cao2014crema,
  author={Houwei Cao and David G. Cooper and Michael K. Keutmann and Ruben C. Gur and Ani Nenkova and Ragini Verma},
  journal={IEEE Transactions on Affective Computing}, 
  title={{CREMA-D: Crowd-Sourced Emotional Multimodal Actors Dataset}}, 
  year={2014},
  volume={5},
  number={4},
  pages={377-390},
  doi={10.1109/TAFFC.2014.2336244}
}

@inproceedings{rousseau2012ted,
  author    = {Rousseau, Anthony and Deléglise, Paul and Estève, Yannick},
  title     = {{TED-LIUM: An Automatic Speech Recognition Dedicated Corpus}},
  booktitle = {Proceedings of the Eighth International Conference on Language Resources and Evaluation ({LREC}`12)},
  pages     = {125--129},
  year      = {2012},
  publisher = {European Language Resources Association (ELRA)}
}

@article{nagrani2020voxceleb,
  author  = {Nagrani, Arsha and Chung, Joon Son and Xie, Weidi and Zisserman, Andrew},
  title   = {{VoxCeleb: Large-Scale Speaker Verification in the Wild}},
  journal = {Computer Speech \& Language},
  volume  = {60},
  pages   = {101027},
  year    = {2020},
  issn = {0885-2308},
  doi = {https://doi.org/10.1016/j.csl.2019.101027},
}

@article{busso2008iemocap,
  author  = {Busso, Carlos and others},
  title   = {{IEMOCAP: Interactive Emotional Dyadic Motion Capture Database}},
  journal = {Journal of Language Resources and Evaluation},
  volume  = {42},
  number  = {4},
  pages   = {335--359},
  year    = {2008},
  doi     = {10.1007/s10579-008-9076-6},
}

@inproceedings{yadav2024ties,
  author    = {Yadav, Prateek and Tam, Derek and Choshen, Leshem and Raffel, Colin A. and Bansal, Mohit},
  title     = {{TIES-Merging: Resolving Interference When Merging Models}},
  booktitle = {Thirty-seventh Conference on Neural Information Processing Systems},
  year      = {2023}
}

@inproceedings{shi2020aishell,
  title     = {{AISHELL-3: A Multi-Speaker Mandarin TTS Corpus}},
  author    = {Shi, Yao and Bu, Hui and Xu, Xin and Zhang, Shaoji and Li, Ming},
  year      = {2021},
  booktitle = {Interspeech 2021},
  pages     = {2756--2760},
  doi       = {10.21437/Interspeech.2021-755},
  issn      = {2958-1796},
}

@inproceedings{lin2024daisy,
  title     = {{DAISY: Data Adaptive Self-Supervised Early Exit for Speech Representation Models}},
  author    = {Lin, Tzu-Quan and Lee, Hung-yi and Tang, Hao},
  year      = {2024},
  booktitle = {Interspeech 2024},
  pages     = {4513--4517},
  doi       = {10.21437/Interspeech.2024-626}
}

@inproceedings{ardila2019common,
  author    = {Ardila, Rosana and others},
  title     = {{Common Voice: A Massively-Multilingual Speech Corpus}},
  booktitle = {Proceedings of the Twelfth Language Resources and Evaluation Conference},
  pages     = {4218--4222},
  year      = {2020},
}

@inproceedings{plantinga2024parameter,
  author    = {Plantinga, Peter and Yoo, Jaekwon and Girma, Abenezer and Dhir, Chandra},
  title     = {{Parameter Averaging Is All You Need To Prevent Forgetting}},
  booktitle = {2024 IEEE Spoken Language Technology Workshop (SLT)},
  pages     = {271--278},
  year      = {2024},
  publisher = {IEEE},
  doi       = {10.1109/SLT61566.2024.10832275}
}

@inproceedings{ramesh2024task,
  author    = {Ramesh, Gowtham and Audhkhasi, Kartik},
  title     = {{Task Vector Algebra for ASR Models}},
  booktitle = {ICASSP 2024 - 2024 IEEE International Conference on Acoustics, Speech and Signal Processing (ICASSP)},
  pages     = {12256--12260},
  year      = {2024},
  publisher = {IEEE},
  doi       = {10.1109/ICASSP48485.2024.10447848}
}

@inproceedings{murata2024attribute,
  author    = {Murata, Masato and Miyazaki, Koichi and Koriyama, Tomoki},
  title     = {{An Attribute Interpolation Method in Speech Synthesis by Model Merging}},
  booktitle = {Interspeech 2024},
  pages     = {3380--3384},
  year      = {2024},
  doi       = {10.21437/Interspeech.2024-208},
}

@article{cheng2024task,
  author  = {Cheng, Yao-Fei and others},
  title   = {{Task Arithmetic for Language Expansion in Speech Translation}},
  journal = {arXiv preprint arXiv:2409.11274},
  year    = {2024}
}

@inproceedings{shi2023multi,
  author    = {Shi, Jiatong and Inaguma, Hirofumi and Ma, Xutai and Kulikov, Ilia and Sun, Anna},
  title     = {{Multi-resolution HuBERT: Multi-resolution Speech Self-Supervised Learning with Masked Unit Prediction}},
  booktitle = {The Twelfth International Conference on Learning Representations},
  year      = {2024},
  pages = {1--19}
}

@article{yang2024large,
  author={Yang, Shu-wen and others},
  journal={IEEE/ACM Transactions on Audio, Speech, and Language Processing}, 
  title={{A Large-Scale Evaluation of Speech Foundation Models}}, 
  year={2024},
  volume={32},
  pages={2884-2899}
}

@incollection{lopes2011phone,
  author = {Lopes, Carla and Perdigao, Fernando},
  title = {{Phoneme Recognition on the TIMIT Database}},
  booktitle = {Speech Technologies},
  publisher = {IntechOpen},
  address = {Rijeka},
  year = {2011},
  editor = {Ivo Ipsic},
  chapter = {14},
  doi = {10.5772/17600}
}

@article{wu2024emo,
  author  = {Wu, Haibin and others},
  title   = {{EMO-SUPERB: An In-Depth Look at Speech Emotion Recognition}},
  journal = {arXiv preprint arXiv:2402.13018},
  year    = {2024}
}

@article{oord2018representation,
  author  = {van den Oord, Aaron and Li, Yazhe and Vinyals, Oriol},
  title   = {{Representation Learning with Contrastive Predictive Coding}},
  journal = {arXiv preprint arXiv:1807.03748},
  year    = {2018}
}

@INPROCEEDINGS{yang2023fast,
  author={Guanrou Yang and Ziyang Ma and Zhisheng Zheng and Yakun Song and Zhikang Niu and Xie Chen},
  booktitle={2023 IEEE Automatic Speech Recognition and Understanding Workshop (ASRU)}, 
  title={{Fast-HuBERT: an Efficient Training Framework for Self-Supervised Speech Representation Learning}}, 
  year={2023},
  pages={1-7},
  publisher = {IEEE},
  doi={10.1109/ASRU57964.2023.10389778}
}

@inproceedings{lin2023spurious,
  author    = {Lin, Yong and Tan, Lu and Hao, Yifan and Wong, Honam and Dong, Hanze and Zhang, Weizhong and others},
  title     = {{Spurious Feature Diversification Improves Out-of-Distribution Generalization}},
  booktitle={The Twelfth International Conference on Learning Representations},
  year      = {2024},
  pages = {1--14}
}

@inproceedings{wang2023task,
  title     = {{Task-Agnostic Structured Pruning of Speech Representation Models}},
  author    = {Wang, Haoyu and Wang, Siyuan and Zhang, Wei-Qiang and Hongbin, Suo and Wan, Yulong},
  year      = {2023},
  booktitle = {Interspeech 2023},
  pages     = {231--235},
  doi       = {10.21437/Interspeech.2023-1442}
}

@inproceedings{shi2023ml,
  title     = {{ML-SUPERB: Multilingual Speech Universal PERformance Benchmark}},
  author    = {Shi, Jiatong and others},
  year      = {2023},
  booktitle = {Interspeech 2023},
  pages     = {884--888},
  doi       = {10.21437/Interspeech.2023-1316}
}

@inproceedings{chang2024colld,
  author={Heng-Jui Changa and Ning Dong and Ruslan Mavlyutov and Sravya Popuri and Yu-An Chung},
  booktitle={ICASSP 2024 - 2024 IEEE International Conference on Acoustics, Speech and Signal Processing (ICASSP)}, 
  title={{COLLD: Contrastive Layer-to-Layer Distillation for Compressing Multilingual Pre-Trained Speech Encoders}}, 
  year={2024},
  pages={10801-10805},
  publisher = {IEEE},
  doi={10.1109/ICASSP48485.2024.10446637}
}

@inproceedings{lai2021semi,
  author    = {Cheng-I Lai and Yung-Sung Chuang and Hung-Yi Lee and Shang-Wen Li and James Glass},
  title     = {{Semi-Supervised Spoken Language Understanding via Self-Supervised Speech and Language Model Pretraining}},
  booktitle = {ICASSP 2021 - 2021 IEEE International Conference on Acoustics, Speech and Signal Processing (ICASSP)},
  pages     = {7468--7472},
  year      = {2021},
  publisher = {IEEE},
  doi       = {10.1109/ICASSP39728.2021.9414922}
}

@inproceedings{hu2022lora,
  author    = {Hu, Edward J. and others},
  title     = {{LoRA: Low-Rank Adaptation of Large Language Models}},
  booktitle = {The Tenth International Conference on Learning Representations},
  year      = {2022},
  pages = {1--13}
}

@inproceedings{chen2023explore,
  author={Zih-Ching Chen and Chin-Lun Fu and Chih-Ying Liu and Shang-Wen Li and Hung-yi Lee},
  booktitle={2022 IEEE Spoken Language Technology Workshop (SLT)}, 
  title={{Exploring Efficient-Tuning Methods in Self-Supervised Speech Models}}, 
  year={2023},
  pages={1120-1127},
  publisher = {IEEE},
  doi={10.1109/SLT54892.2023.10023274}
}

@article{xu2019forget,
  title={{Forget Me Not: Reducing Catastrophic Forgetting for Domain Adaptation in Reading Comprehension}},
  author={Ying Xu and Xu Zhong and Antonio Jimeno-Yepes and Jey Han Lau},
  journal={2020 International Joint Conference on Neural Networks (IJCNN)},
  year={2019},
  pages={1--8}
}

@inproceedings{chen-etal-2020-recall,
    title={{Recall and Learn: Fine-tuning Deep Pretrained Language Models with Less Forgetting}},
    author={Chen, Sanyuan and others},
    booktitle={Proceedings of the 2020 Conference on Empirical Methods in Natural Language Processing (EMNLP)},
    year={2020},
    publisher={Association for Computational Linguistics},
    doi={10.18653/v1/2020.emnlp-main.634},
    pages={7870--7881},
}

@article{raghavan2024engineering,
  title={Engineering flexible machine learning systems by traversing functionally invariant paths},
  author={Raghavan, Guruprasad and Tharwat, Bahey and Hari, Surya Narayanan and Satani, Dhruvil and Thomson, Matt},
  journal={Nature Machine Intelligence},
  volume={6},
  number={10},
  pages={1179--1196},
  year={2024},
  publisher={Nature Publishing Group},
  doi={10.1038/s42256-024-00902-x}
}

@inproceedings{kahn2020self,
  title={Self-training for end-to-end speech recognition},
  author={Kahn, Jacob and Lee, Ann and Hannun, Awni},
  booktitle={ICASSP 2020 - 2020 IEEE International Conference on Acoustics, Speech and Signal Processing (ICASSP)},
  pages={7084--7088},
  year={2020},
  organization={IEEE}
}

@inproceedings{li2024training,
  title={{Training-free model merging for multi-target domain adaptation}},
  author={Wenyi Li and Huan-ang Gao, Mingju Gao and Beiwen Tian and Rong Zhi and Hao Zhao},
  booktitle={European Conference on Computer Vision (ECCV)},
  pages={419--438},
  year={2024},
  organization={Springer}
}

@inproceedings{marczak2024magmax,
  title={{MagMax: Leveraging Model Merging for Seamless Continual Learning}},
  author={Marczak, Daniel and Twardowski, Bart{\l}omiej and Trzci{\'n}ski, Tomasz and Cygert, Sebastian},
  booktitle={European Conference on Computer Vision (ECCV)},
  pages={379--395},
  year={2024},
  organization={Springer}
}

@inproceedings{zhou2024metagpt,
  title={{MetaGPT: Merging Large Language Models Using Model Exclusive Task Arithmetic}},
  author={Zhou, Yuyan and Song, Liang and Wang, Bingning and Chen, Weipeng},
  booktitle={Proceedings of the 2024 Conference on Empirical Methods in Natural Language Processing (EMNLP)},
  year={2024},
  pages={1711--1724},
  publisher={Association for Computational Linguistics}
}

@inproceedings{yang2024finetuning,
  author={Yang, Hejung and Kang, Hong-Goo},
  booktitle={ICASSP 2024 - 2024 IEEE International Conference on Acoustics, Speech and Signal Processing (ICASSP)}, 
  title={{On Fine-Tuning Pre-Trained Speech Models With EMA-Target Self-Supervised Loss}}, 
  year={2024},
  pages={6360-6364},
  organization={IEEE}
}

@inproceedings{chen2023emotionfinetune,
  author={Chen, Li-Wei and Rudnicky, Alexander},
  booktitle={ICASSP 2023 - 2023 IEEE International Conference on Acoustics, Speech and Signal Processing (ICASSP)}, 
  title={{Exploring Wav2vec 2.0 Fine Tuning for Improved Speech Emotion Recognition}}, 
  year={2023},
  pages={1-5},
  organization={IEEE}
}

@inproceedings{pan24c_interspeech,
 title     = {{Attentive Merging of Hidden Embeddings from Pre-trained Speech Model for Anti-spoofing Detection}},
 author    = {Pan, Zihan and Liu, Tianchi and Sailor, Hardik B. and Wang, Qiongqiong},
 year      = {2024},
 booktitle = {Interspeech 2024},
 pages     = {2090--2094},
 doi       = {10.21437/Interspeech.2024-1472},
 issn      = {2958-1796},
}

@inproceedings{guragain2024speech,
 title={{Speech foundation model ensembles for the controlled singing voice deepfake detection (ctrsvdd) challenge 2024}},
 author={Guragain, Anmol and Liu, Tianchi and Pan, Zihan and Sailor, Hardik B and Wang, Qiongqiong},
 booktitle={2024 IEEE Spoken Language Technology Workshop (SLT)},
 pages={774--781},
 year={2024},
 organization={IEEE}
}

@inproceedings{pasad2021layer,
  title={{Layer-wise analysis of a self-supervised speech representation model}},
  author={Pasad, Ankita and Chou, Ju-Chieh and Livescu, Karen},
  booktitle={2021 IEEE Automatic Speech Recognition and Understanding Workshop (ASRU)},
  pages={914--921},
  year={2021},
  organization={IEEE}
}

@article{pratap2020mls,
  title={{MLS: A Large-Scale Multilingual Dataset for Speech Research}},
  author={Pratap, Vineel and Xu, Qiantong and Sriram, Anuroop and Synnaeve, Gabriel and Collobert, Ronan},
  journal={arXiv preprint arXiv:2012.03411},
  year={2020}
}

@inproceedings{liu2024dora,
  title={{DoRA: Weight-Decomposed Low-Rank Adaptation}},
  author={Liu, Shih-Yang and Wang, Chien-Yi and Yin, Hongxu and Molchanov, Pavlo and Wang, Yu-Chiang Frank and Cheng, Kwang-Ting and Chen, Min-Hung},
  booktitle={Proceedings of the 41st International Conference on Machine Learning},
  year={2024},
  pages={{32100}--{32121}}
}

@inproceedings{zhou2023going,
  title={{Going Beyond Linear Mode Connectivity: The Layerwise Linear Feature Connectivity}},
  author={Zhou, Zhanpeng and Yang, Yongyi and Yang, Xiaojiang and Yan, Junchi and Hu, Wei},
  booktitle={The Thirty-seventh Annual Conference on Neural Information Processing Systems},
  volume={36},
  pages={60853--60877},
  year={2023}
}

@article{schonemann1966procrustes,
  title={A generalized solution of the orthogonal Procrustes problem},
  author={Sch{\"o}nemann, Peter H.},
  journal={Psychometrika},
  volume={31},
  number={1},
  pages={1--10},
  year={1966},
  publisher={Springer}
}

@article{hoerl1970ridge,
  title={Ridge regression: Biased estimation for nonorthogonal problems},
  author={Hoerl, Arthur E and Kennard, Robert W},
  journal={Technometrics},
  volume={12},
  number={1},
  pages={55--67},
  year={1970},
  publisher={Taylor \& Francis}
}

\newpage

\vfill

\end{document}